\pdfoutput=1
\documentclass[table]{article}

\usepackage[preprint]{coling}

\usepackage{times}
\usepackage{latexsym}

\usepackage[T1]{fontenc}
\usepackage[utf8]{inputenc}

\usepackage{microtype}

\usepackage{inconsolata}

\usepackage{xcolor}

\usepackage{graphicx}
\usepackage{amsmath}
\usepackage{hyperref}       % hyperlinks
\usepackage{url}            % simple URL typesetting
\usepackage{booktabs}       % professional-quality tables
\usepackage{amsfonts}       % blackboard math symbols
\usepackage{nicefrac}       % compact symbols for 1/2, etc.
\usepackage{microtype}      % microtypography
\usepackage{algorithm}
\usepackage{algpseudocode}
\usepackage{multirow}  % for multirow% for colors in tables
\usepackage{amsmath}
\usepackage{amssymb}
\usepackage{subcaption}
\usepackage{mathptmx} % Times New Roman font
\usepackage{mathtools}

\usepackage{hyperref}
\usepackage{url}
\usepackage{float}

\usepackage{bm}        % For bold math symbols

\newcommand{\mbx}{\mathbf{x}}      % Bold x
\newcommand{\btheta}{\boldsymbol{\theta}}  % Bold theta using boldsymbol
\newcommand{\mbK}{\mathbf{K}}      % Bold K
\newtheorem{theorem}{Theorem}

\title{Propulsion: Steering LLM with Tiny Fine-Tuning}

\author{
Md Kowsher\textsuperscript{1}, Nusrat Jahan Prottasha\textsuperscript{1}, Prakash Bhat\textsuperscript{2,$\ddagger$} \\
\textsuperscript{1}University of Central Florida, FL, USA, \textsuperscript{2}Amazon, USA \\
\texttt{\{md.kowsher, nusrat.prottasha\}@ucf.edu, bhatprak@amazon.com}
}

\begin{document}
\maketitle
\begin{abstract}

The rapid advancements in Large Language Models (LLMs) have revolutionized natural language processing (NLP) and adjacent fields, yet fine-tuning these models for specific tasks remains computationally expensive and risks degrading pre-learned features. To address these challenges, we propose \emph{Propulsion}, a novel parameter-efficient fine-tuning (PEFT) method designed to optimize task-specific performance while drastically reducing computational overhead. Inspired by the concept of controlled adjustments in physical motion, \emph{Propulsion} selectively re-scales specific dimensions of a pre-trained model, guiding output predictions toward task objectives without modifying the model’s parameters. By introducing lightweight, trainable \emph{Propulsion} parameters at the pre-trained layer, we minimize the number of parameters updated during fine-tuning, thus preventing the overfitting or overwriting of existing knowledge. Our theoretical analysis, supported by Neural Tangent Kernel (NTK) theory, shows that \emph{Propulsion} approximates the performance of full fine-tuning with far fewer trainable parameters. Empirically, \emph{Propulsion} reduces the parameter count from 355.3 million to a mere 0.086 million—achieving over a 10x reduction compared to standard approaches like LoRA—while maintaining competitive performance across benchmarks. Code is available at: \textcolor{red}{\href{https://github.com/Kowsher/Propulsion}{\texttt{https://github.com/Kowsher/Propulsion}}}.

\end{abstract}
\renewcommand*{\thefootnote}{\fnsymbol{footnote}}
\footnotetext[3]{This work does not relate to Prakash's position at Amazon.}
\renewcommand*{\thefootnote}{\arabic{footnote}}

\begin{figure*}[!t]
%\begin{wrapfigure}{r}{0.7\textwidth}
%\captionsetup{font=footnotesize}

 \begin{center}
    \includegraphics[width=0.95\linewidth]{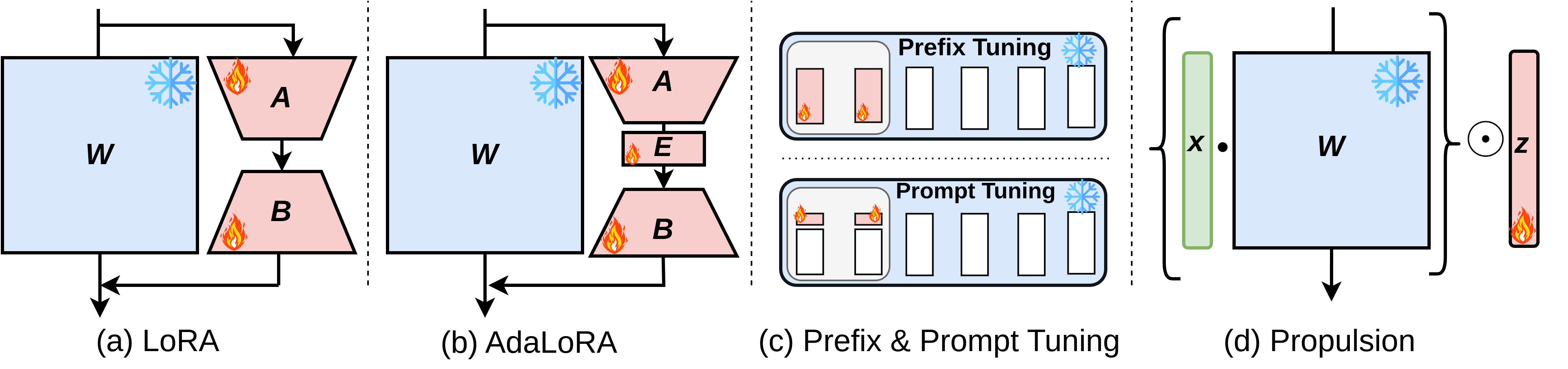}

   \end{center}
%\end{wrapfigure}
\caption{A detailed illustration of the model architectures for five different adapters: (a) LoRA, (b) AdaLoRA, (c) Prefix \& Prompt Tuning, and (d) Propulsion. In the diagrams, \textbf{\textit{W}} represents the pre-trained weight matrix, which is kept frozen, while \textbf{\textit{X}} denotes the input. The matrices \textbf{\textit{A}}, \textbf{\textit{B}}, and \textbf{\textit{E}} are trainable and of lower rank. The variable \textbf{{z}} indicates the \emph{Propulsion} parameter.}
\label{fig:Propulsion_cmp}
\end{figure*}

\section{Introduction}

Training large language models consumes significant computational resources, sometimes taking up to six months~\cite{zhao2023survey}. This creates bottlenecks in AI development and raises environmental concerns~\cite{rillig2023risks}. To mitigate this, we often fine-tune pre-trained models like BERT~\cite{devlin2018bert}, GPT~\cite{mann2020language}, and RoBERTa~\cite{liu2019roberta} instead of training from scratch. However, fine-tuning these pre-trained models is still challenging due to their large sizes; for instance, modern LLMs can have up to $7$~billion parameters~\cite{jiang2023mistral, touvron2023llama, almazrouei2023falcon, le2023bloom}. Traditional full model fine-tuning is effective but often too expensive and inefficient, limited by computational resources and time~\cite{bender2021dangers, kim2024memory, wu2024continual}.

Recent advances have explored the realm of PEFT~\cite{xu2023parameter, kowsher2023tuning} techniques as a solution to these challenges. Methods such as adapter layers~\cite{lin-etal-2020-exploring, houlsby2019parameter}, prompt tuning~\cite{lester2021power}, low-rank adaptation~\cite{hu2021lora}, quantization~\cite{gray1998quantization}, selective row or columns tuning \cite{kowsher2024rocoft}, and lightweight fine-tuning~\cite{liu2021enabling} alternatives propose modifications that require adjusting only a fraction of the model's total parameters. These approaches, while promising, often involve trade-offs between efficiency, performance, and adaptability, thus there is still room to improve the combined utility. 
To address the limitations of existing PEFT methods, we introduce \emph{Propulsion}: a novel approach for fine-tuning that leverages the observation that small, targeted changes in the output vectors of a model's layers can lead to substantial shifts in the model's overall behavior.  In physical dynamics, propulsion can steer or change an object’s trajectory through small, controlled bursts of force ~\cite{turchi1998Propulsion, budashko2020thrusters}. Similarly, our \emph{Propulsion} method applies minimal yet strategic adjustments or re-scaling to the pre-trained dimensions of a neural network, as effectively "steering" the model’s responses towards desired outcomes with minimal energy expenditure and maximal retention of pre-learned features. To do this, we introduce a series of trainable linear parameters—denoted as "\emph{Propulsion} parameters". These parameters are finely tuned to amplify or attenuate specific aspects of the model's behavior, thereby optimizing performance on specific tasks with minimal computational overhead. Figure~\ref{fig:Propulsion_cmp} compares the different PEFT methods with our \emph{Propulsion} approach.

To support our method theoretically, we analyze \emph{Propulsion} in the context of the NTK framework. The NTK, introduced by ~\citet{jacot2018neural}, characterizes the training dynamics of neural networks in the regime where the width of the network tends to infinity. Under this framework, it has been shown that fine-tuning methods such as LoRA approximate the full fine-tuning of neural networks by focusing on a low-rank subspace~\cite{jang2024lora, tomihari2024understanding}. Similarly, our analysis demonstrates that \emph{Propulsion} closely approximates the NTK of full fine-tuning by updating only a diagonal subset of the model’s parameters. This theoretical grounding ensures that \emph{Propulsion} achieves similar performance to full fine-tuning, despite its significantly reduced computational requirements.

We evaluate the effectiveness of our approach across several benchmarks on different language models. Our experimental results show that \emph{Propulsion} outperforms current PEFT techniques while requiring fewer trainable parameters. For instance, \emph{Propulsion} uses about $12$ times fewer parameters than AdaLoRA and achieves higher accuracy (details in Section \ref{sec:exp}).

\section{Propulsion}\label{sec:Propulsion}

We introduce a clear outline of the \emph{Propulsion} concept and its practical benefits. Consider that we have a pre-trained language model $\mathbb{M}$ with $N$ layers, such as \( L = \{L_{1}, L_{2}, \ldots, L_{N}\} \), where we freeze all parameters. We represent any given input as \( x \in \mathbb{R}^{s \times d_{in}} \), where \( s \) denotes the sequence length of tokens,  \( d \) represents the dimension of each token, and $x$ can be any hidden layer's output or input of next following layer of the neural networks, Key, Queries, and Values, and so on. Given \(x\) as the input, we extract \( V_i = L_{i}(x; W) \in \mathbb{R}^{s \times d_{out}} \) with pre-trained frozen weight $W \in \mathbb{R}^{d_{in} \times d_{out}}$.

To introduce task-specific modifications, we initialize a trainable \emph{Propulsion} matrix \( \mathcal{Z} \in \mathbb{R}^{N \times d_{out}} \), where \( \mathbf{z_i} = \{z_{1}, z_{ 2}, \ldots, z_{d_{out}}  \}\in \mathcal{{Z}} \). Each \( \mathbf{z_i} \) performs an element-wise scalar transformation to each corresponding element \( \mathbf{\mathbf{v_j}} \in V_i \) to steer the output projection of \( L_{i} \), where \(  \mathbf{\mathbf{v_j}} = \{v_{1}, v_{2} \ldots v_{d_{out}}\}\) represents the $j-th$ token representation of output $V_i$ from layer \(L_{i}\).

We train $\mathbf{z_i}$ by calculating the element-wise multiplication $ \mathbf{\mathbf{v_j}} \odot \mathbf{z_i}$ to generate $\mathbf{v_j}^\prime$, where \( \odot \) denotes the element-wise multiplication operation performed between \(z_{d_{out}}\) and every element \( v_{d_{out}}\) within the output vector \( {\mathbf{v_j}} \). We can define this operation as : 
\begin{equation}
    \mathbf{v_j}^{\prime} = [v_{1} \cdot z_1,v_{2} \cdot z_2,...,v_{d_{out}} \cdot z_{d_{out}}]
    \label{eq:base_Propulsion}
\end{equation} 

Similarly, by following Equation~\ref{eq:base_Propulsion}; for all $s$ tokens, we can steer the output of $V_i$ by training the \emph{Propulsion} $\mathbf{z_i}$, which can be defined as :

\begin{equation}
    V_i^{\prime} = [\mathbf{v_1} \odot \mathbf{z_i},\mathbf{v_2} \odot \mathbf{z_i},...,\mathbf{v_s} \odot \mathbf{z_i}]
    \label{eq:Propulsion}
\end{equation} 

% \Ella{Is the transformation an element-wise multiplication between \(z_{i}\) and \(v_{i}\), or is it a Kronecker product?}g
% \Nusrat{we use this notation for element wise multification. If you have better notation you can use}

Once \(V_i^{\prime}\) has been calculated, it is used as the next input to extract the output of the next layer. So the transformed output \(V_i^{\prime}\) of layer \(L_{i}\) is used as the input \(x\) to layer \(L_{i+1}\)

We enhance the \emph{Propulsion} concept by incorporating polynomial scaling to the \emph{Propulsion} parameter \( \mathbf{z_i} \). By raising \( \mathbf{z_i} \) to the power of \( k \), termed as polynomial scaling, we allow for a more flexible and dynamic adjustment of the model's responses to input features. This scaling adjusts the magnitude of the propulsion effect, providing a method to vary the influence of the propulsion parameters across different stages of learning or different parts of the data. We can define this operation as :
\begin{equation}
    V_i^{\prime} = [\mathbf{v_1} \odot \mathbf{z_i}^k, \mathbf{v_2} \odot \mathbf{z_i}^k, \ldots, \mathbf{v_s} \odot \mathbf{z_i}^k]
\end{equation}

% Similarly for all layers $N$, we can use a trainable \emph{Propulsion} matrix $\mathcal{Z} \in \mathbb{R}^{N \times d}$ to steer every layer output. 

\begin{figure}[!t]
%\begin{wrapfigure}{r}{0.7\textwidth}
%\captionsetup{font=footnotesize}

\label{fig:method}
 \begin{center}
    \includegraphics[width=0.80\linewidth]{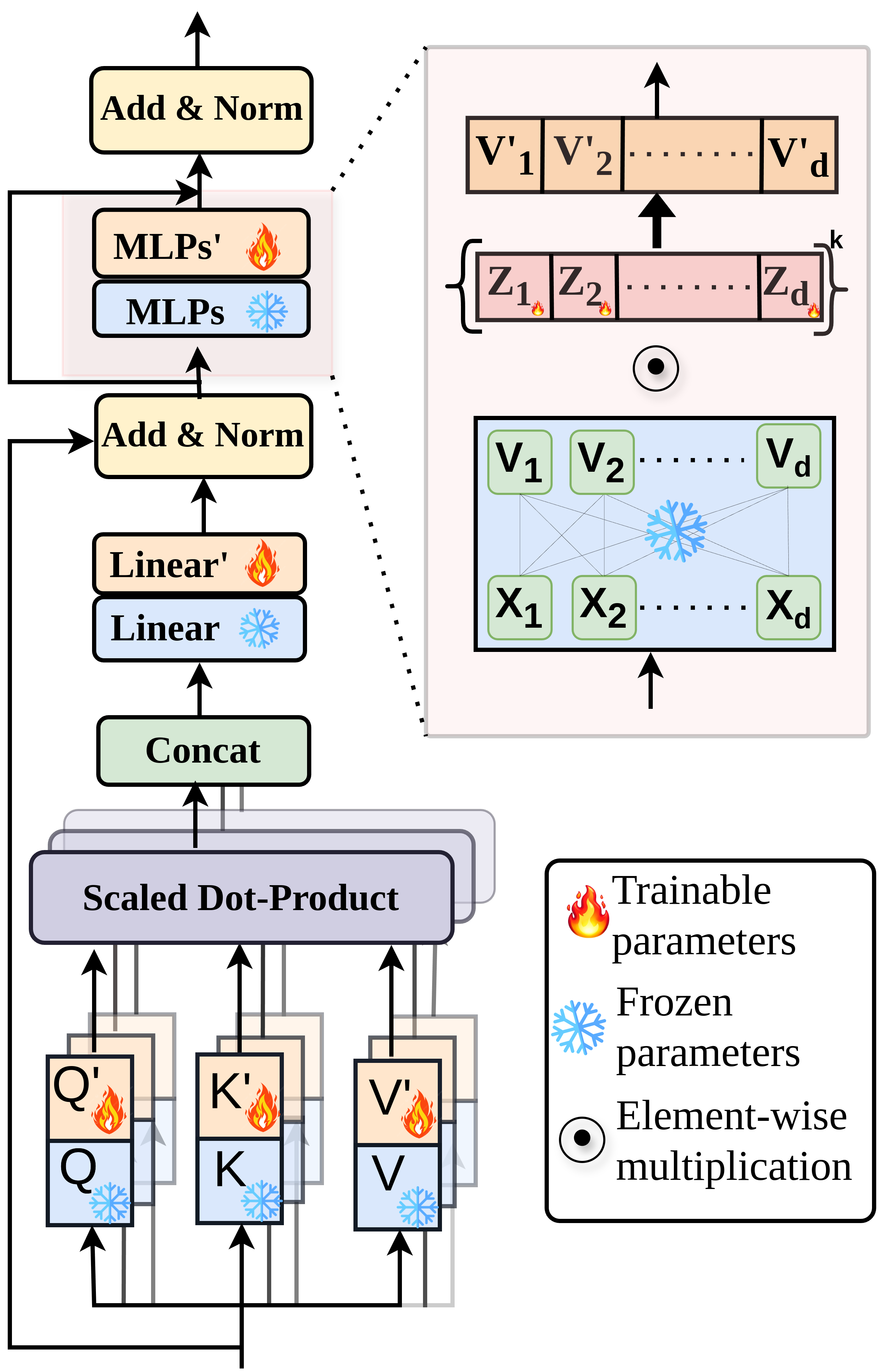}

   \end{center}
%\end{wrapfigure}
\caption{\emph{Propulsion} in Transformer Block.  Within the figure, the red cells represent trainable parameters while the blue cells represent the frozen parameters. The \emph{Propulsion} layers above shows where our method executes during model fine-tuning.  All layers use the same \emph{Propulsion} matrix, but are modified by their corresponding vector \(z_{i}\).}\label{fig:Propulsion}
\end{figure}

In Figure~\ref{fig:Propulsion}, we illustrate the general structure of our \emph{Propulsion} method in the Transformer block that modifies the output of K, Q, V, and MLP matrix through element-wise multiplication with \emph{Propulsion} trainable parameters to fine-tune the LLMs efficiently.

\section{Neural Tangent Kernel (NTK) Analysis}
The NTK, introduced by \citet{jacot2018neural}, characterizes how small changes in a network’s parameters affect its output. In the NTK regime, where the width of the network becomes very large, the training dynamics of neural networks are determined by the NTK, which remains nearly constant during training \cite{afzalcan}.

In this section, we analyze the Propulsion method in the NTK regime and show that the NTK of Propulsion approximates the NTK of full fine-tuning.

\begin{theorem}
\label{thm:propulsion_ntk}
Let $\phi_P(\mathbf{x}; \theta_t)$ be the output of the Propulsion model at time step $t$, where the base matrix $\theta_0$ is pre-trained and fixed, and the Propulsion matrix $\mathbf{z}_t$ is updated during training. Let $\phi_F(\mathbf{x}; \theta_t)$ be the output of the fully fine-tuned model at time step $t$.  Under the NTK regime, where the width $d$ of the network is sufficiently large, the NTK for Propulsion fine-tuning approximates the NTK for full fine-tuning with high probability. Formally, for inputs $\mathbf{x}, \mathbf{x}' \in \mathbb{R}^d$, the NTK for Propulsion satisfies:
\[
   \mbK^F\left(\mbx, \mbx^{\prime}\right)  \approx \mbK^P\left(\theta_0\mbx_i, \theta_0\mbx_j\right)
\]
Furthermore, the error between the NTK for Propulsion and the NTK for full fine-tuning can be bounded using the Johnson-Lindenstrauss Lemma. Specifically, for any $\epsilon > 0$ and constant $c$, with high probability:
\small
\[
\Pr\left[\left| (\theta_0\mbx_i)^\top(\theta_0\mbx_j) - \mathbf{x_i}^\top \mathbf{x_j} \right|  \right] \geq 1 - 4 \exp\left(- \frac{(\epsilon^2 - \epsilon^3) d}{4}\right)
\]
\normalsize
\end{theorem}
The full theoretical proof of this theorem is provided in \textbf{Appendix \ref{thm:proof}}. Additionally, in \textbf{Appendix \ref{sec:NTK_results}}, we present the empirical results supporting this theory, and in \textbf{Appendix \ref{sec:kernel_behaviour}}, we provide a detailed analysis of the NTK regime of Propulsion.

\begin{table*}[htbp]
\centering

\scalebox{.70}{
\begin{tabular}{l|c|c|c|c|c|c|c|c|c|c|c}
\hline
\textbf{Model} & \textbf{PEFT Method} & \textbf{\#TPs} & \textbf{CoLA} & \textbf{SST2} & \textbf{MRPC} & \textbf{STS-B} & \textbf{QQP} & \textbf{MNLI} & \textbf{QNLI} & \textbf{RTE} & \textbf{Avg.} \\
\hline

% Here's where we make RoB span across 3 rows
\multirow{10}{*}{RoB\textsubscript{B}} & FT & 124.6M & 59.84 & 92.89 & 85.24/88.18 & 90.48/90.16 & 90.18/87.02 & 86.27 & 91.17 & 72.43 & 83.56 /88.45 \\ \cline{2-12}\cline{2-12}
& Adapter\textsuperscript{S} & 7.41M & 60.32 & 92.14 & 89.24/ \underline{85.29} & 90.25/\underline{90.09} & \textbf{90.81/86.55} & \textbf{87.33} & 90.84 & 73.56 & 84.31/\underline{87.31}\\
& Prompt tuning & 0.61M & 49.37 & 91.09 & 74.83/72.72 & 82.44/83.11 & 82.99/78.35 & 80.57 & 80.03 & 58.12 & 74.93/78.06\\
& Prefix-tuning & 0.96M & 55.31 & 92.17 & 87.25/83.24 & 88.48/88.32 & 87.75/84.09 & 85.21 & 90.77 & 54.51 & 80.18/85.21 \\
& (IA)\textsuperscript{3} & 0.66M & 59.58 & 92.02 & 87.00/82.52 & 90.30/90.32 & 87.99/84.10 & 83.95 & 90.88 & 71.12 & 82.85/85.64\\
& BitFit & 0.086M & 61.38 & 92.67 & 88.22/84.41 & 90.34/90.27 & 88.12/84.11 & 84.64 & 91.09 & \underline{75.58} & 84.20/86.26 \\
%& Child-Tuning\textsubscript{D} & - & 60.33 & 93.58 & 89.22/92.20 & 91.14/90.93 & 90.98/88.04 & 87.40 & 92.20 & 77.62 & 85.31/85.29 \\
& LoRA & 0.89M & 60.09 & 92.40 & 88.50/84.68 & 90.66/90.83 & 88.83/85.21 & 86.54 & 92.02 & 72.92 & 83.99/86.90 \\
& AdaLoRA & 1.03M & 59.82 & 91.69 & 88.99/85.03 & 90.83/90.73 & 88.58/84.98 & 86.26 & 91.43 & 70.04 & 83.45/86.91\\
& MAM Adapter & 46.78M & 58.42 & \textbf{93.19} & \underline{89.31} /85.21 & 90.74/90.42 & 88.31/83.20 & 86.63 & 90.19 & 72.62 & 83.67/86.27 \\
& PROPETL \textsubscript{Adapter} & 1.87M & \textbf{63.11} & 92.18 & 85.25/81.82 & \underline{91.33} /\textbf{91.04} & 89.22/85.79 & 86.49 & \underline{92.56} & 75.54 & \underline{84.46}/86.21 \\
& PROPETL \textsubscript{Prefix} & 10.49M & 60.18 & 91.36 & 86.73/84.98 & 90.30/90.19 & 88.54/85.05 & 86.22 & 91.51 & 63.31 & 82.26/86.74\\
& PROPETL \textsubscript{LoRA} & 1.77M & 61.72 & 92.54 & 87.42/83.87 & 90.76/90.55 & 88.90/85.55 & \underline{86.84} & 92.04 & 67.39 & 83.45/86.65\\ 

\cline{2-12}
& \cellcolor{lightgray!33.333}Propulsion(All) & \cellcolor{lightgray!33.333}\underline{0.086M} & \cellcolor{lightgray!33.333}\underline{61.76} & \cellcolor{lightgray!33.333}\underline{93.18} & \cellcolor{lightgray!33.333}\textbf{89.34}/ \textbf{85.99} & \cellcolor{lightgray!33.333}\textbf{91.37}/\underline{90.92} & \cellcolor{lightgray!33.333}89.11/\underline{86.53} & \cellcolor{lightgray!33.333}86.41 & \cellcolor{lightgray!33.333}\textbf{92.79} & \cellcolor{lightgray!33.333}\textbf{75.66} & \cellcolor{lightgray!33.333}\textbf{84.95}/\textbf{87.81}\\

& \cellcolor{lightgray!33.333}Propulsion(Attn) & \cellcolor{lightgray!33.333}\textbf{0.028M} & \cellcolor{lightgray!33.333}58.51 & \cellcolor{lightgray!33.333}92.03 & \cellcolor{lightgray!33.333}89.01/85.14 & \cellcolor{lightgray!33.333}89.36/89.96 & \cellcolor{lightgray!33.333}86.73/84.80 & \cellcolor{lightgray!33.333}85.13 & \cellcolor{lightgray!33.333}89.89 & \cellcolor{lightgray!33.333}75.02 & \cellcolor{lightgray!33.333}83.21/86.63\\

\hline
\multirow{10}{*}{RoB\textsubscript{L}} & FT & 355.3M & 65.78 & 95.50 & 92.22/94.28 & 91.74/91.96 & 90.83/88.68 & 89.21 & 93.19 & 81.40 & 87.48/91.64 \\ \cline{2-12}
& Adapter\textsuperscript{S} & 19.77M & 62.03 & 94.65 & 90.19/87.94 & \underline{92.58}/92.42 & \underline{92.19/88.50} & \underline{91.00} & 94.31 & 81.25 & 87.27/\underline{89.62} \\ 
& Prompt-tuning & 1.07M & 60.22 & 93.61 & 79.04/76.29 & 78.51/78.99 & 80.74/75.16 & 68.15 & 89.13 & 60.29 & 76.21/76.81\\
& Prefix-tuning & 2.03M & 59.01 & 93.76 & 88.24/86.37 & 90.92/91.07 & 88.88/85.45 & 89.30 & 93.32 & 74.01 & 84.68/87.63 \\
& (IA)\textsuperscript{3} & 1.22M & 60.17 & 94.61 & 90.52/87.33 & 92.22/86.25 & 89.45/86.25 & 88.63 & 94.25 & 81.23 & 86.38/86.61 \\
& Bitfit& 0.225M & \textbf{66.72} & 95.10 & \underline{90.70/88.38} & 91.93/\textbf{93.38} & 89.48/86.43 & \underline{89.98} & 94.47 & \underline{85.73} & 88.01/89.39 \\
%& Child-Tuning\textsubscript{D} & - & 63.08 & 95.07 & 90.69/93.43 & 92.36/92.18 & 91.52/88.75 & 35.45 & 93.15 & 86.25 & 80.95/80.92 \\
& LoRA & 1.84M & 64.47 & \textbf{95.67} & 90.50/86.19 & 91.66/91.44 & 90.15/86.91 & 90.76 & 95.00 & 79.78 & 87.24/88.18 \\
& AdaLoRA & 2.23M & \underline{65.85} & 94.95 & \textbf{91.46}/87.34 & 92.05/91.80 & 89.60/86.30 & 90.36 & 94.62 & 77.98 & 88.20/88.48 \\
& MAM Adapter & 122.20M & 64.39 & 95.08 & 90.12/87.77 & 92.44/92.18 & 90.87/86.65 & 90.62 & 94.31 & 86.62 & 88.05/88.86 \\
& PROPETL \textsubscript{Adapter} & 5.40M & 65.55 & 94.82 & 89.71/86.54 & 91.92/91.67 & 90.67/87.74 & \textbf{91.37} & \underline{95.20} & \textbf{85.89} & \underline{88.14}/88.65 \\
& PROPETL \textsubscript{Prefix} & 26.85M & 62.24 & 94.17 & 90.04/87.92 & 90.70/90.49 & 89.30/86.30 & 90.33 & 94.73 & 79.71 & 86.40/88.23 \\
& PROPETL \textsubscript{LoRA} & 4.19M & 61.90 & 94.93 & 89.06/86.19 & 91.66/91.38 & 90.93/88.05 & 90.53 & 94.93 & 82.57 & 87.06/88.54\\\cline{2-12}

& \cellcolor{lightgray!33.333}Propulsion(All) & \cellcolor{lightgray!33.333}\underline{0.225M} & \cellcolor{lightgray!33.333}64.53 & \cellcolor{lightgray!33.333}\underline{95.10} & \cellcolor{lightgray!33.333} 90.47/\textbf{88.85} & \cellcolor{lightgray!33.333}\textbf{92.78}/\underline{92.58} & \cellcolor{lightgray!33.333}\textbf{92.26/88.91} & \cellcolor{lightgray!33.333}90.52 & \cellcolor{lightgray!33.333}\textbf{95.34} & \cellcolor{lightgray!33.333}85.30 & \cellcolor{lightgray!33.333}\textbf{88.28/90.11} \\

& \cellcolor{lightgray!33.333}Propulsion(Attn) & \cellcolor{lightgray!33.333}\textbf{0.073M} & \cellcolor{lightgray!33.333}62.31 & \cellcolor{lightgray!33.333}94.02 & \cellcolor{lightgray!33.333}89.78/87.95 & \cellcolor{lightgray!33.333}90.16/90.86 & \cellcolor{lightgray!33.333}88.02/86.19& \cellcolor{lightgray!33.333}89.54 & \cellcolor{lightgray!33.333}94.00 & \cellcolor{lightgray!33.333}83.07 & \cellcolor{lightgray!33.333}86.36/88.33
 \\
\hline
\end{tabular}
}
\caption {Performance Comparison of RoBERTa Models on GLUE Tasks: Metrics include MCC for CoLA, Accuracy for SST-2, Accuracy/F1-score for MRPC and QQP, Pearson/Spearman correlation for STS-B, and Accuracy for MNLI, QNLI, and RTE. "Propulsion(All)" applies Propulsion to all layers (Embedding, MLP, Attention), while "Propulsion(Attn)" applies it only to the Attention layer. Propulsion(All)\textsuperscript{3} refers to three Propulsion mechanisms in each layer.}

\label{table: result-1}
\end{table*}

\begin{table*}[ht]
\centering
\scalebox{.78}{
\begin{tabular}{l|ccc|ccc}
\hline
\textbf{PEFT Method} & \textbf{\#TPs} & \textbf{SQuADv1.1} & \textbf{SQuADv2.0} & \textbf{\#TPs} & \textbf{XSum} & \textbf{CNN/DailyMail}  \\
\hline
FT & 460M & 82.83 / 88.14 & 82.92 / 83.75 & 460M &  40.73 / 16.19 / 30.13 & 39.16 / 18.92 / 37.04  \\ \hline

Prompt tuning & 0.155M & 74.52 / 78.42 & 73.59 / 76.72 & 0.524M & 38.24 / 14.46 / 27.89 &  37.42 / 17.43 / 34.92 \\
Prefix-tuning & 2.683M & 78.38 / 82.94 & 74.94 / 79.04 &4.482M & 38.24 / 15.16 / 28.84 &  38.32 / 17.72 / 35.76 \\
LoKr & 0.089M & 80.64 / 86.45 & 80.14 / 81.96 & 0.194M 
& 39.03 / 16.14 / 30.42 &  39.12 / 17.98 / 37.75 \\
Bitfit & 0.161M & 80.53 / 86.25 & 79.06 / 83.75 & 0.885M & 39.10 / 16.87 / 30.43 & 39.93 / 18.12 / 38.85 \\
LoHa & 0.885M & 81.43 / 88.02 & 81.67 / 85.01 & 1.769M &39.12 / 17.08 / 31.39 &  39.98 / 18.84 / 38.01\\
LoRA & 0.442M & 81.64 / 87.16 & \textbf{82.76} / 85.75 & 1.763M & 40.63 / \textbf{18.44} / \textbf{32.35} & \textbf{40.74} / \underline{19.10} / \underline{39.24}\\
AdaLoRA & 0.663M &  \underline{81.16} / \underline{87.75} & 82.63 / \textbf{85.82} & 2.655M & \underline{40.95} / \underline{18.28} / \underline{31.84} & 40.53 / 18.24 / \textbf{39.63} \\ \hline

\hline
\cellcolor{lightgray!33.333} Propulsion(All)& \cellcolor{lightgray!33.333}0.161M & \cellcolor{lightgray!33.333}\textbf{81.73 / 88.07} & \cellcolor{lightgray!33.333} \underline{82.68}/ \underline{85.81} & \cellcolor{lightgray!33.333}0.330M & \cellcolor{lightgray!33.333}\textbf{40.98} / {18.18} / {31.42} & \cellcolor{lightgray!33.333}\underline{40.56} / \textbf{19.28} / 38.76\\

\cellcolor{lightgray!33.333}Propulsion(Attn) & \cellcolor{lightgray!33.333}0.055M & \cellcolor{lightgray!33.333}80.95 / 87.20 & \cellcolor{lightgray!33.333}81.02 / 85.50 & \cellcolor{lightgray!33.333}0.110M & \cellcolor{lightgray!33.333} 38.64 / 15.45 / 29.25 & \cellcolor{lightgray!33.333}38.74 / 17.08 / 35.03\\ 
\hline
\end{tabular}
}
\caption{Performance of DeBERTaV3-base and BART-large on SQuAD v1.1 and v2.0 benchmarks with EM/F1 and ROUGE scores (ROUGE-1/ROUGE-2/ROUGE-L). Here, the \textbf{bolded} values indicate the best performance, while the \underline{underlined} values represent the second-best performance.}

\label{table: result-2}
\end{table*}

\section{Experiments} \label{sec:exp}

We evaluate our methods on NLP tasks, including the General Language Understanding Evaluation (GLUE) benchmark, question answering, text summarization, common sense reasoning, and arithmetic reasoning. The details of the training and algorithm are described in Appendix~\ref{sec:training}.

\subsection{Baselines}

We use well-known PEFT methods for our baseline comparisons, including Adapter \cite{houlsby2019parameter}, Prompt Tuning \cite{lester2021power}, Prefix-Tuning \cite{li2021prefix}, (IA)\textsuperscript{3} \cite{liu2022few}, Bitfit \cite{zaken2021bitfit}, LoRA \cite{hu2021lora}, AdaLoRA \cite{zhang2023adaptive}, MAM Adapter \cite{he2021towards}, PROPETL \cite{zeng2023one}, LoKr \cite{edalati2022krona}, and LoHa \cite{hyeon2021fedpara}. The implementations used for these methods come from the Hugging Face \cite{peft}. The experimental setup follows that of \citet{xu2023parameter} for the GLUE benchmark; for the question answering and text summarizing datasets, we have followed \citet{zhang2023adaptive}.

\subsection{Language Model Performance}
\textbf{Datasets : } For the GLUE Benchmark, we evaluate our \emph{Propulsion} method on CoLA, SST-2, MRPC, STS-B, QQP, MNLI, QNLI, and RTE tasks of the GLUE Benchmarks\cite{wang2018glue}.
We also use SQuAD v1.1 \cite{rajpurkar2016squad} and SQuAD v2.0 \cite{rajpurkar2018know} datasets to measure performance on question-answering tasks, and we use the XSum \cite{narayan2018don} and CNN/DailyMail \cite{hermann2015teaching} datasets to measure text summarization performance. 

{\textbf{Model Selection \& Hyperparameter : }} For the GLUE benchmark, the models we select for fine-tuning are RoBERTa-base (RoB\textsubscript{B}) with 125M parameters and RoBERTa-large (RoB\textsubscript{L}) with 355M parameters from \citet{liu2019roberta}. We set the \emph{Propulsion} degree to $15$ as discussed in Section \ref{sec:Propulsion} for SST-2, QQP, RTE, and STS-B; $55$ for QNLI and MRPC; and $20$ for the other GLUE datasets.

For the SQuAD v1.1 and SQuAD v2.0 datasets, we employ DeBERTaV3-base~\cite{he2020deberta}. For both SQuAD v1.1 and SQuAD v2.0, we set the \emph{Propulsion} degree to $35$. 

For the XSum and CNN/DailyMail datasets, we chose the BART-large model \cite{lewis2019bart} with 406M parameters. 
For XSum and CNN/DailyMail, we set the Propulsion degrees to $35$ and $25$.

\textbf{Results : } Table \ref{table: result-1} shows the GLUE task validation results of \emph{Propulsion}, in comparison with baselines, we can see that \emph{Propulsion} can achieve better or on-par performance compared with existing PEFT approaches on the GLUE dataset but with much less trainable parameters. Overall, \emph{Propulsion} exhibits enhancements of $2.48\%$, $3.15\%$, and $3.17\%$ in accuracy over AdaLoRA, PROPETL Prefix, and (IA)\textsuperscript{3}, respectively, and $1.94\%$, $1.87\%$, and $8.92\%$ improvements in the F1 score.

Table \ref{table: result-2} compares the validation performance of \emph{Propulsion} and other PEFT methods on question-answering and text summarization tasks.  For question answering tasks, \emph{Propulsion} outperforms the other PEFT methods on both the SQuAD datasets.  \emph{Propulsion} beats AdaLoRA, the second highest performing PEFT method, by 0.66 in EM and 0.51 in F1 score while being 7.89 times smaller in parameter size.  Comparing to LoKr, which has the least number of trainable parameters amongst the baseline PEFT methods, \emph{Propulsion} outperforms LoKr by 2.92 in EM and by 2.29 in F1-score while having fewer parameters.

For text summarization, \emph{Propulsion} has the highest ROUGE-1 score among the baseline PEFT methods on both datasets.  It also has the best ROUGE-2 score and the second-best ROUGE-L score on the CNN/DailyMail dataset. For XSum, LoRA and AdaLoRA have higher ROUGE-2 and ROUGE-L scores than \emph{Propulsion}. This may be due to the limitations \emph{Propulsion} may have by being constrained by the model's dimension, whereas LoRA and AdaLoRA have more flexibility with more parameters, which is evident by higher ROUGE-2/L scores. Despite this, \emph{Propulsion's} performance on CNN/DailyMail shows that it achieves on-par performance with methods like LoRA and AdaLoRA while having a significantly smaller parameter size. For both tables, we used the validation set to test the performance.

\begin{table*}[htbp]
\centering
\scalebox{.6}{
\begin{tabular}{l|c|cccccccc|ccccc}
\hline
\textbf{LLM} & \textbf{Method} & \textbf{BoolQ} & \textbf{PIQA} & \textbf{SIQA} & \textbf{H.Swag} & \textbf{W.Grande} & \textbf{ARC-e} & \textbf{ARC-c} & \textbf{OBQA} & \textbf{MultiArith} & \textbf{GSM8K} & \textbf{AddSub} & \textbf{SingleEq} & \textbf{SVAMP}\\
\hline
\multirow{5}{*}{\textbf{BLOOMz$_{7B}$}} & Prefix & 58.53 & 62.24 & 65.41 & 48.32 & 66.63 & 68.13 & 49.32 & 63.51 & 78.41 & 66.45 & 67.52 & 66.94 & 49.10\\
& AdaLoRA & 64.94 & \textbf{74.68} & 72.49 & 52.89 & 68.30 & \underline{73.21} & 56.59 & \underline{72.85} & \underline{79.43} & 70.25 & 68.93  & \underline{70.93} & \underline{53.89} \\
& Parallel & 63.30 & 73.33 & 71.01 & 52.50 & 71.60 & 69.45 & 54.14 & 68.60 & 78.90 & 70.17 & 70.33 & 70.84 & 53.95 \\
& LoRA & \underline{65.89} & \underline{73.92} & \underline{73.33} & \underline{56.65} & \underline{71.39} & \textbf{73.46} & \underline{57.15} & 72.31 & \textbf{79.50} & \underline{70.93} & \underline{70.90} & 70.59 & 53.85 \\
& Propulsion & \textbf{66.38} & 74.63 & \textbf{74.62} & \textbf{57.25} & \textbf{72.33} & 73.09 & \textbf{57.61} & \textbf{73.12} & 79.36 & \textbf{70.95} & \textbf{70.92} &  \textbf{71.22} & \textbf{54.52}\\
\hline

\multirow{5}{*}{\textbf{GPT-J$_{6B}$}} & Prefix & 62.28 & 65.04 & 67.72 & 44.15 & 63.71 & 63.59 & 46.47 & 58.31 & 83.12 & 67.44 & 75.25 & 78.46 & 49.12\\
& AdaLoRA & 65.19 & 67.58 & \textbf{71.22} & 45.16 & \textbf{66.03} & \underline{64.10} & \textbf{47.75} & \underline{63.92} & 88.51 & \underline{72.45} & 80.21 & \textbf{82.03} & 56.14 \\
& Parallel & 63.17 & \underline{67.91} & 68.97 & 45.79 & 66.06 & 62.42 & 45.32 & 60.42 & \underline{89.11} & 72.04 & 80.50  & \underline{81.50} & 55.43 \\
& LoRA & \underline{65.50} & 67.63 & 69.46 & \underline{45.60} & \underline{66.37} & 63.56 & 46.81 & 63.82 & 88.30 & 72.22 & \underline{80.60} & 81.24 & \underline{56.63} \\
& Propulsion & \textbf{65.97} & \textbf{68.05} & \underline{69.96} & \textbf{45.99} & 66.18 & \textbf{64.45} &  \underline{46.95} & \textbf{64.56} & \textbf{89.19} & \textbf{72.82} & \textbf{81.41} & 81.42 & \textbf{56.68}\\
\hline

\multirow{5}{*}{\textbf{LLaMA$_{7B}$}} & Prefix & \underline{67.33} & \underline{79.46} & 75.80 & 70.04 & 72.11 & 71.67 & 57.33 & 69.98  & 84.18 & 68.47 & 81.04 & 80.00 & 52.17\\
& AdaLoRA & 67.03 & 78.69 & 76.06 & 75.85 & 76.47 & 76.26 & 60.36 & 74.22 & 89.81 & \underline{77.07} & \underline{86.70} & \underline{83.01} & \underline{60.25} \\
& Parallel & 65.02 & 78.10 & \textbf{77.52} & 75.57 & 76.78 & 75.48 & 60.54 & 74.02 & \underline{90.20} & 76.13 & 86.55 & \textbf{83.70} & 59.16\\
& LoRA & 67.09 & 79.37 & 76.15 & \textbf{76.86} & \textbf{77.54} & \underline{76.54} & \underline{60.55} & \underline{74.63} & 90.13 & 75.68 & 84.67  & 82.14 & 59.94\\
& Propulsion & \textbf{68.99} & \textbf{79.47} & \underline{77.02} & \underline{76.73} & \underline{77.06} & \textbf{76.64} & \textbf{61.29} & \textbf{74.76} & \textbf{90.21} & \textbf{77.57} & \textbf{87.63} & 82.60 & \textbf{60.51} \\
\hline

\multirow{4}{*}{\textbf{LLaMA$_{13B}$}} & Prefix & 68.38 & 80.99 & 77.80 & 75.00 & 76.35 & 77.62 & 61.32 & 72.94 & 87.22 & 71.09 & 84.09  & 81.28 & 58.25 \\
& AdaLoRA & \underline{71.71} & 82.55 & 78.88 & 90.60 & 83.01 & 83.04 & \underline{67.33} & \textbf{81.76} & 90.55 & \textbf{80.19} & 87.00 & \underline{87.10} & \underline{66.03} \\
& Parallel & 71.39 & 83.33 & 78.32 & \textbf{91.40} & \underline{83.24} & 83.34 & 66.43 & 80.99 & \underline{90.88} & \underline{79.24}  & \textbf{88.16} & 87.08 & 65.63 \\
& LoRA & 71.19 & \underline{83.99} & \textbf{79.15} & \underline{90.86} & \underline{83.24} & 83.35 & 67.05 & 81.37  & 90.27 & 78.90  & 86.89 & 86.07 & 65.85 \\
& Propulsion & \textbf{71.93} & \textbf{84.12} & \underline{79.01} & 90.73 & \textbf{83.60} & \textbf{83.44} & \underline{67.64} & \underline{81.38} & \textbf{90.91} & 78.71 & \underline{87.64} & \textbf{87.11}& \textbf{66.67} \\
\hline
\end{tabular}
}
\caption{Accuracy comparison of Commonsense and Mathematical reasoning performance across different PEFTs with 3\% performance reduction.}
\label{table:example}
\end{table*}

\begin{figure*}[!t]
    \centering
    \includegraphics[width=0.90\linewidth]{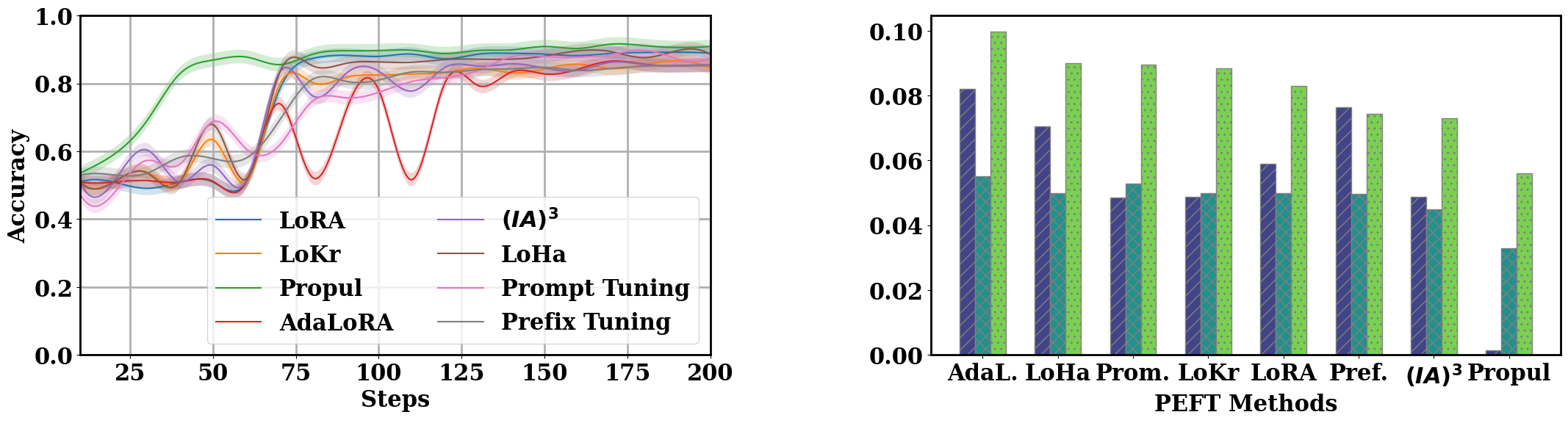}
  
    \label{fig:compare_params}
\caption{Comparative Analysis of PEFT Methods on the SST-2 Dataset. On the right-side graph, we shortened the following method names: AdaLoRA to AdaL., Prompt Tuning to Prom., Propulsion to Propul, and Prefix-Tuning to Pref.  In this graph, purple represents the percentage of parameters after applying these methods, the cyan represents the total training time in hours, and the green represents the iteration time in seconds.}
\label{fig:compare1}
\end{figure*}

\begin{figure}[!t]
%\begin{wrapfigure}{r}{0.7\textwidth}
%\captionsetup{font=footnotesize}
\centering

 % \begin{center}
      % \includegraphics[width=1.05\linewidth]{Memory Cost.png}
    
      \includegraphics[width=0.95\linewidth]{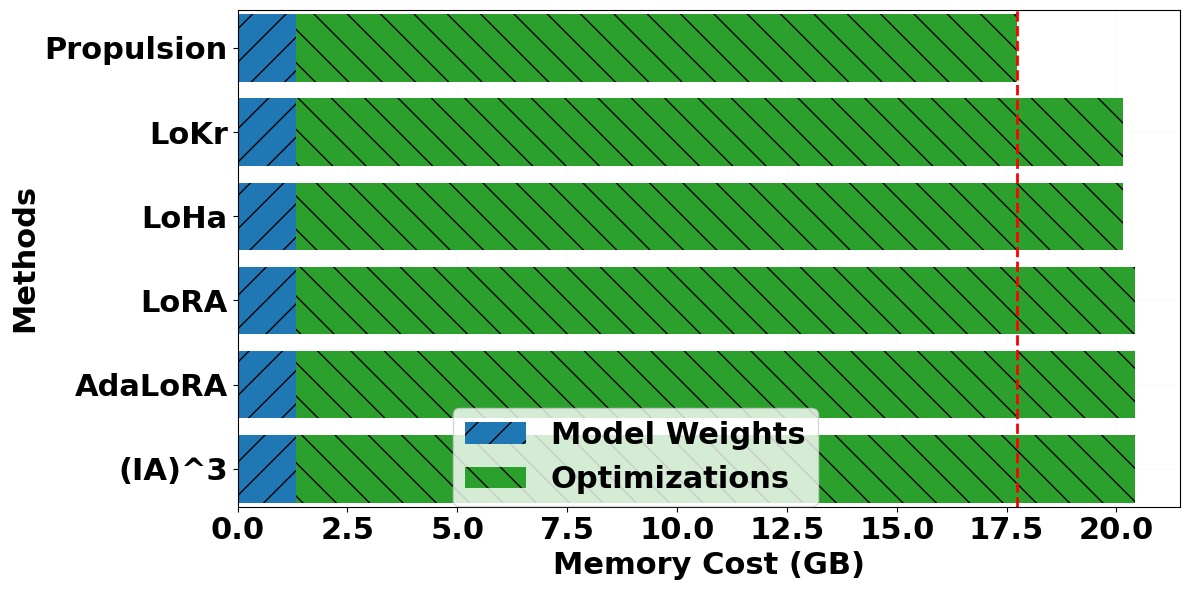}
   % \end{center}
%\end{wrapfigure}
\caption{Memory Cost Comparison of PEFT Methods. The blue bars represent the memory cost of the original model weights, whereas the green bars represent the optimization memory cost for each of these methods.}\label{fig:memory}
\end{figure}

\subsection{Large Language Models Performance}

\textbf{Datasets : } We perform a thorough evaluation using thirteen benchmark datasets, covering common sense reasoning and mathematical reasoning tasks.

For common sense reasoning, we employ a diverse range of datasets, including BoolQ \cite{clark2019boolq}, PIQA \cite{bisk2020piqa}, SIQA \cite{sap2019socialiqa}, HellaSwag \cite{zellers2019hellaswag}, WinoGrande \cite{sakaguchi2021winogrande}, ARC-easy  and  ARC-challenge \cite{clark2018think}, and OBQA \cite{OpenBookQA2018}, to ensure a comprehensive assessment of our model's ability to handle various facts of common sense reasoning.

For arithmetic reasoning tasks, we also use several professional datasets, including MultiArith \cite{roy2016solving}, GSM8K \cite{cobbe2021training}, AddSub \cite{hosseini2014learning}, SingleEq \cite{koncel2015parsing}, and SVAMP \cite{patel2021nlp} to evaluate the performance of our model on solving  various different arithmetic reasoning-related problems. 

\textbf{Model Selection \& Hyperparameters:} For the commonsense and mathematical reasoning tasks described, we select several LLMs to fine-tune using both standard baselines and our proposed \emph{Propulsion} methods for comparison.

The LLMs chosen include BLOOMz (7B parameters) \cite{muennighoff2022crosslingual}, GPT-J (6B parameters), LLaMA (7B parameters, denoted as \textbf{LLaMA\textsubscript{7B}}), and LLaMA (13B parameters, denoted as \textbf{LLaMA\textsubscript{13B}}). For BLOOMz, \textbf{LLaMA\textsubscript{7B}}, \textbf{LLaMA\textsubscript{13B}}, and \textbf{GPT-J\textsubscript{6B}}, we set the \emph{Propulsion} degree to 15 for both reasoning tasks. Additionally, we apply a dropout rate of 0.1 for both hidden layers and attention mechanisms, along with L2 regularization. Each model layer is fine-tuned using 5 distinct \emph{Propulsion} parameters to assess the effectiveness of our approach.

\begin{table}[ht]
    \centering
   
\scalebox{0.72}{
    \begin{tabular}{l|l|l|l}
    \hline
        \textbf{Methods} & \textbf{Space}  & \textbf{Time} & \textbf{\#TPs} \\
        \hline
        \emph{Propulsion} & \(O(d)\)  & \(O(d)\) & $d$ \\
        \hline
        FT & $O(d \times d)$ & $O(d \times d)$ &  $d^2$ \\
        $(IA)^3$ & $O(d_{k}+d_{v}+d_{ff})$ & $O(d_{k}+d_{v}+d_{ff})$ & $3d$\\
        %\hline
        Prompt  & \(O(d \times l_{p})\) & \(O(d \times l_{p})\) & $l_p.d$\\
        %\hline
        Prefix & \(O(L \times d \times l_{p})\) & \(O(L \times d \times l_{p})\) & $L.l_p.d$\\ %\hline
    
        LoRA & $O((d+d) \times r)$ & $O((d+d) \times r)$ & $2dr$\\
        LoRA-FA & $O((d+d) \times r)$ & $O((d+d) \times r)$ & $dr$\\

        AdaLoRA & $O((d+d+r) \times r)$ & $O((d+d+r) \times r)$ & $2dr+r^2$\\
        LoHA & $O(2r  \times (d+d))$ & $O(2r  \times (d+d))$ & $4dr$\\
        \hline
       
\end{tabular}}
 \caption{Space/Time Complexity and Total Trainable Parameters (\#TPs) for \emph{Propulsion} method and baseline methods for single layer $W\in \mathbb{R}^{d \times d}$. Within this table, we define \(d_{k}, d_{v},\) and \(d_{ff}\) as the dimensions of three learned vectors in \((IA)^{3}\); and \(l_{p}\) as the length of the prompt added to the input/layers in prompt tuning and prefix-tuning. For LoRA-type methods, we use \(r\) to represent the rank dimensions.}\label{table:complexity}

\end{table}
\begin{figure*}[!t]

    \centering
    % \begin{minipage}[t]{.488\textwidth}
    %     \centering
        % \includegraphics[width=0.999\linewidth]{degrees.pk.png} 

    \includegraphics[width=0.9\linewidth]{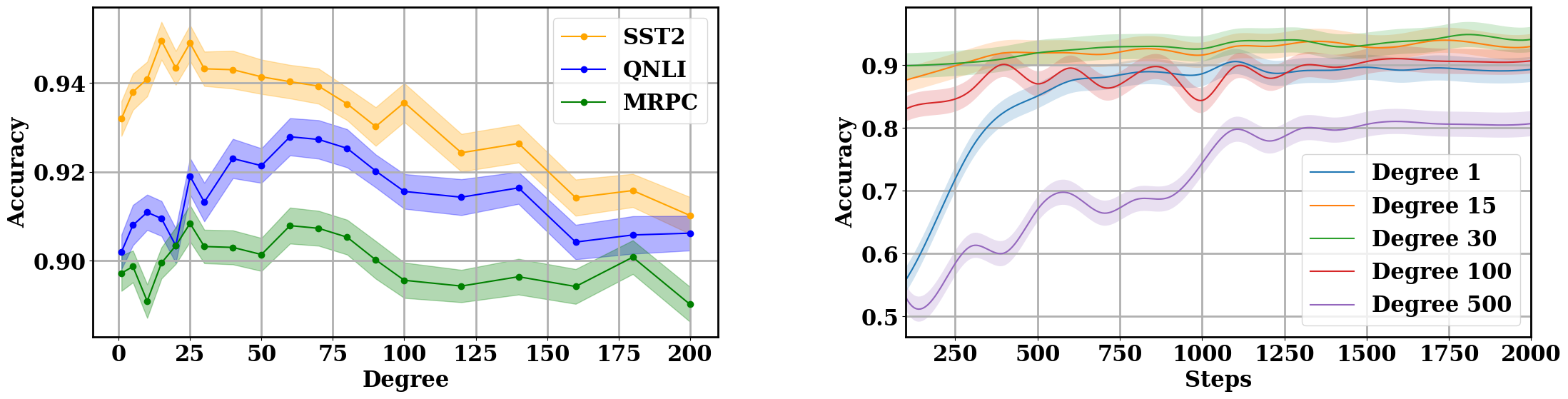}
  
    \label{fig:compare_sub1}
    % \end{minipage}%
    % \hfill
    % \begin{minipage}[t]{.512\textwidth}
    %     \centering
        % \includegraphics[width=0.999\linewidth]{PushingValueStudy.png}

        % \label{fig:compare_sub12}
    % \end{minipage}
\caption{ Left: performance vs. degree for SST-2, QNLI, and MRPC. Right: training steps vs. accuracy for SST-2.}
\label{fig:compare_degree}
\end{figure*} 
\textbf{Results : } Table \ref{table:example} shows the accuracy results on all four LLMs across the thirteen benchmarks. Across the board, \emph{Propulsion} outperforms state-of-the-art PEFT methods on both commonsense and mathematical reasoning tasks. On average across the four LLMs tested on these benchmarks, \emph{Propulsion} shows competitive performance on all of the benchmarks while maintaining the highest accuracy on benchmarks like GSM8K.Notably, fine-tuning the BLOOMz and GPT-J models demonstrates competitive performance against the baseline methods. For datasets like SIQA, and  HellaSwag, our method achieves $1.33\%$, $0.97\%$ improvement than the state-of-the-art PEFT method on accuracy. And for LLaMA model fine-tuning (\textbf{LLaMA\textsubscript{7B}}, \textbf{LLaMA\textsubscript{13B}}), \emph{Propulsion} also reach better performance than other baselines on most datasets, e.g. on the AddSub and SVQMP datasets, \emph{Propulsion} shows enhancements of $0.97\%$ and $0.66\%$ in accuracy over the state-of-the-art PEFT method.
While maintaining or improving accuracy, \emph{Propulsion} also has a much smaller percentage of total parameters.
Additional experiments on LLMs are described in Appendix~\ref{sec:llmmore}.

\subsection{Efficiency Comparison} Our study evaluates diverse PEFT techniques on their performance, training efficiency, and memory usage. We conduct these experiments using the SST-2 dataset, divided into $64$ batches. We train on a H100 with $80$ GB of memory. The default parameters include the learning rate of  \(1 \times 10^{-4}\), the weight decay of $0.02$, dropout $0.1$.

\textbf{Training efficiency:} Figure \ref{fig:compare1} illustrates the training convergence of our models and baselines. On the left side of the figure, it shows that our \emph{Propulsion} model exhibits a fast convergence and achieves a higher accuracy of $0.9$ in just $50$ iterations, whereas the baseline AdaLoRA method requires approximately $200$ iterations to attain an accuracy of $0.87$, and the LoRA method requires almost $75$ iterations to reach an accuracy comparable to that of \emph{Propulsion}. Furthermore, the other methods, including LoKr, (IA)\textsuperscript{3}, and LoHa, require more than $150$ iterations to achieve an accuracy of $0.8$.

\textbf{Parameter Efficiency : }
In terms of parameters, we present the efficiency of each method in Tables \ref{table: result-1} and \ref{table: result-2}, as well as a graphical representation in Figure \ref{fig:compare1} (Right). It is clear that \emph{Propulsion} demonstrates superior efficiency in terms of faster training time, and reduced memory usage because of its parameter reduction. Table \ref{table:complexity} compares the space/time complexities and total trainable parameters of our \emph{Propulsion} method to other baseline PEFT methods.

\textbf{Memory Efficiency:} In terms of memory efficiency of the GPU, as illustrated in Figure \ref{fig:memory}, \emph{Propulsion} consumes only approximately $17.2$ GB of GPU memory for training, including model weights and optimization. In comparison, other baseline methods consume more than $20.0$ GB of GPU memory, making \emph{Propulsion} approximately $1.5$ times more memory-efficient than other PEFT methods.
Additionally, in Appendix~\ref{sec:delta_cmp} (Table~\ref{tab:delta_cmp}), we present a comparison of delta weight reparameterization methods for the backward pass during optimization

\subsection{Ablation Study}

\textbf{Propulsion Degree Initialization:} In this section, we explore the impact of the \emph{Propulsion} degree as a hyperparameter on model performance across different datasets. Figure~\ref{fig:compare_degree} (left) shows the accuracy on SST-2, QNLI, and MRPC for degrees ranging from 0 to 200. SST-2 achieves its highest accuracy of 95\% at a degree of 25, while QNLI peaks at 94\% between 50 and 75 degrees, and MRPC at 92\% around 25 degrees. After reaching peak accuracy, both QNLI and MRPC show a decline, indicating overfitting as the \emph{Propulsion} degree increases.

Figure~\ref{fig:compare_degree} (right) shows the training dynamics on SST-2. Lower degrees (1 and 15) converge faster, achieving high accuracy early, while higher degrees (100 and 500) take longer. By 2000 steps, all degrees converge, but lower degrees stabilize faster, suggesting they are more effective for rapid learning, with higher degrees needing more steps for similar performance.

\begin{table}[t]
\centering

\scalebox{0.70}{
\begin{tabular}{c|cccccc}
\toprule
\textbf{Layer}  & \textbf{SST-2} & \textbf{MRPC} & \textbf{QQP} & \textbf{QNLI} & \textbf{RTE} & \textbf{Params} \\
\hline
Embedding & 73.45 & 70.32 & 75.28 & 79.38 & 66.3 & 0.0115M \\
MLP & 92.42 & 86.42 & 82.38 & 89.35 & 72.43 & 0.0115M \\
Key & 92.52 & 86.84 & 83.95 & 88.14 & 72.19 & 0.0115M \\
Value & 92.68 & 86.59 & 83.05 & 88.76 & 73.93 & 0.0115M \\
Query & 91.53 & 86.99 & 83.84 & 89.28 & 73.68 & 0.0115M \\
K+Q+V & 92.72 & 89.01 & 85.82 & 89.89 & 75.02 & 0.0283M \\
All & 93.18 & 89.34 & 89.11 & 92.79 & 75.66 & 0.0861M \\
\bottomrule
\end{tabular}
}

\caption{Accuracy [\%] and the parameter size for different layer configurations with \emph{Propulsion} across datasets.}
\label{tab:Propulsion_study_transposed}
\end{table}
\textbf{Positional Impact of Propulsion: } Table \ref{tab:Propulsion_study_transposed} shows an ablation analysis of \emph{Propulsion} configured across various layers, including embedding, MLP, Key (K), Query(Q), Value (V), and different combinations of layers. Adding \emph{Propulsion} to the attention mechanism (K + V + Q) achieved an accuracy of $93.72\%$ on the SST-2 dataset. When examined individually, we obtained accuracies of $91.52\%$, $92.52\%$, and $92.68\%$ in the Query, and Value, respectively. However, \emph{Propulsion} in the embedding layer does not yield performance comparable to that of the other layers. Nonetheless, \emph{Propulsion} in all layers leads to substantial accuracy improvements of $94.89\%$, $90.52\%$, $90.86\%$, $92.79\%$, and $77.60\%$  for the SST-2, MRPC, QQP, QNLI, and RTE datasets, respectively.

Additional ablation studies are described in Appendix~\ref{sec:moreabls}.

\section{Related Work}

The development of parameter-efficient fine-tuning (PEFT) techniques is essential in NLP due to the increasing complexity of LLMs. These techniques enhance performance while reducing computational and memory requirements, as demonstrated by recent studies \citep{liu2022few, nguyen2023efficient, chow2024performance}. PEFT techniques have been proven effective across a wide range of NLP tasks, including \citep{fu2023effectiveness, he2021towards}. Previous research \citep{liu2021p, liu2023gpt, zhang2023adaptive, hu2021lora, li2021prefix, zaken2021bitfit} has shown that PEFT techniques can significantly improve the performance of LLMs while utilizing low resources.

Prompt Tuning entails adding learnable parameters as virtual tokens at the model's input \cite{lester2021power} or within each layer \cite{li2021prefix}. Recent advancements have refined these methods for NLU \cite{liu2021p} and NLG \cite{an2022input}, including adding residual connections for stability \cite{razdaibiedina2023residual} and adapting to continual learning \cite{razdaibiedina2023progressive}. Innovative techniques like MixPAVE \cite{yang2023mixpave} and E2VPT \cite{han20232vpt} integrate input and value prompts to boost performance. These methods have significantly enhanced specific NLP tasks such as text classification, machine translation, and dialogue generation.

Low-Rank Adaptation (LoRA), introduced by \citet{hu2021lora}, is a memory-efficient fine-tuning technique extensively studied. \citet{renduchintala2023tied}, \citet{sheng2023s}, and \citet{xia2024chain} explored its multitask learning potential. \citet{wang2023multilora} showed practical applications, while \citet{dettmers2024qlora} optimized memory usage. \citet{lialin2023relora} proposed ReLoRA, requiring a full-rank warm-up. Adaptive methods by \citet{zhang2023adaptive} dynamically adjust low-rank parameters. \citet{edalati2022krona} introduced the Low-Rank Kronecker Product (LoKr), and \citet{shi2024reslora} developed ResLoRA with residual paths. \citet{hyeon2021fedpara} presented the Low-Rank Hadamard Product (LoHa), while \citet{qiu2024controlling} and \citet{liu2024parameterefficient} introduced Orthogonal Finetuning (OFT) and OFT with butterfly factorization (BOFT), using orthogonal matrices to modify pre-trained weights, enhancing fine-tuning efficiency and performance.

Unlike previous PEFT approaches, we propose a new concept of adaptive \emph{Propulsion} that changes the output direction of the model by Propulsion a force to achieve task-specific goals. We adjust the \emph{Propulsion} parameter during the training process, which decides how much push needs to change the direction. (More details related work in Appendix~\ref{sec:Rw_more})

\section{Conclusion}

Fine-tuning extensive language models can be costly in terms of hardware and storage switching expenses, and the financial investment required to host separate instances of diverse tasks is often substantial. We propose \emph{Propulsion}, a parameter-efficient fine-tuning method that adds trainable \emph{Propulsion} parameters to each layer while keeping the original parameters frozen. The goal of \emph{Propulsion} is to achieve task-specific objectives without modifying the original parameters of the LLMs. Our experiments on natural language processing, question answering, text summarization, common sense reasoning, and mathematical reasoning show that \emph{Propulsion} outperforms existing methods in terms of accuracy, efficiency, faster convergence, reduced training time, and lower memory usage. Our results demonstrate that \emph{Propulsion} outperforms current PEFT techniques while requiring fewer trainable parameters. For example, \emph{Propulsion} uses $37$ times fewer parameters than AdaLoRA and achieves $4.05\%$ higher accuracy.

\section{Limitations}
The \emph{Propulsion} method has a few limitations. First, it offers limited control over the model compared to other methods such as LoRA, which allows adjustments through changes in rank. In \emph{Propulsion}, the ability to steer a model is constrained by the number of dimensions in each layer. Essentially, we can only adjust the \emph{Propulsion} parameters equal to the number of dimensions of a layer, which restricts the extent to which we can tweak the model's behavior. Additionally, since each parameter in the \emph{Propulsion} method works independently without influencing others, it may be harder to make coordinated changes across the model. Moreover, the success of  \emph{Propulsion} depends on the quality of the pre-trained language model.

\bibliography{custom}

\appendix

\section{Training Dynamics of Propulsion Explained by NTK}
\label{thm:proof}

\textbf{Theorem 1} Let $\phi_P(\mathbf{x}; \btheta_t)$ be the output of the Propulsion model at time step $t$, where the base matrix $\btheta_0$ is pre-trained and fixed, and the diagonal matrix $\mathbf{Z}_t$ is updated during training. Let $\phi_F(\mathbf{x}; \btheta_t)$ be the output of the fully fine-tuned model at time step $t$. 

Under the NTK regime, where the width $d$ of the network is sufficiently large, the NTK for Propulsion fine-tuning approximates the NTK for full fine-tuning with high probability. Formally, for inputs $\mathbf{x}, \mathbf{x}' \in \mathbb{R}^d$, the NTK for Propulsion satisfies:
\[
   \mbK^F\left(\mbx, \mbx^{\prime}\right)  \approx \mbK^P\left(\btheta_0\mbx_i, \btheta_0\mbx_j\right)
\]
Furthermore, the error between the NTK for Propulsion and the NTK for full fine-tuning can be bounded using the Johnson-Lindenstrauss Lemma. Specifically, for any $\epsilon > 0$ and constant $c$, with high probability:
\small
\[
\Pr\left[\left| (\btheta_0\mbx_i)^\top (\btheta_0\mbx_j) - \mathbf{x_i}^\top \mathbf{x_j} \right|  \right] \geq 1 - 4 \exp\left(- \frac{(\epsilon^2 - \epsilon^3) d}{4}\right)
\]
\normalsize

To establish the proof of the theorem, we first introduce the definitions of the NTK Kernel and the Kernel Behavior specific to the Propulsion method.

\textbf{Definition-1} (NTK Kernel):
Let \( \mathbf{K}(\mbx, \mbx') \) represent the Neural Tangent Kernel (NTK) of a model. The kernel is defined as the inner product of the gradients of the model outputs with respect to the parameters \( \btheta \). Formally, for inputs \( \mbx, \mbx' \in \mathbb{R}^d \), the kernel is given by:
\[
\mathbf{K}(\mbx, \mbx') = \nabla_{\btheta} \phi_P(\mbx; \btheta)^{\top} \nabla_{\btheta} \phi_P(\mbx'; \btheta),
\]
where \( \nabla_{\btheta} \phi_P(\mbx; \btheta) \) represents the gradient of the model output \( \phi_P(\mbx; \btheta) \) with respect to the parameters \( \btheta \).

\textbf{Definition-2} (Kernel Behavior):
Let \( \btheta_t \) represent the parameters of a model at time step \( t \), and let \( \mbx \) be an arbitrary fixed input. The Propulsion model exhibits \textit{kernel behavior} if the following properties are satisfied:

\begin{enumerate}
    \item \textbf{Linearization}: The change in the model's output can be well-approximated by the first-order Taylor expansion. Specifically:
    \small
    \[
    \phi_P(\mbx; \btheta_t) - \phi_P(\mbx; \btheta_{t-1}) \approx \left\langle \nabla \phi_P(\mbx; \btheta_{t-1}), \btheta_t - \btheta_{t-1} \right\rangle,
    \]
    \normalsize
    where \( \nabla \phi_P(\mbx; \btheta) \) is the gradient of the model's output with respect to the parameters \( \btheta \).

    \item \textbf{Fixed Features}: The gradient of the model at time step \( t \) is approximately the same as the gradient at initialization, i.e.,
    \[
    \nabla \phi_P(\mbx; \btheta_t) \approx \nabla \phi_P(\mbx; \btheta_0),
    \]
    where \( \btheta_0 \) refers to the parameters at initialization.
\end{enumerate}

\textbf{Proof:} Let $\btheta_t$ represent the parameters of the network at time step $t$, and $\phi_{\btheta}$ denote the output of the pre-trained network. Under the NTK approximation, the change in the network's output can be expressed as a first-order Taylor expansion:
\[
\phi_{\btheta_{t+1}}(\mathbf{x}) \approx \phi_{\btheta_t}(\mathbf{x}) + \left\langle \nabla_{\btheta_t} \phi_{\btheta_t}(\mathbf{x}), \btheta_{t+1} - \btheta_t \right\rangle.
\]

In this work, we aim to analyze the Propulsion fine-tuning method in the context of NTK, and show that the NTK of Propulsion closely approximates the NTK of full fine-tuning.

\textbf{Kernel Behavior:}
In stochastic gradient descent (SGD), the update to the parameters at step $t$ is given by:
\begin{align}
\btheta_{t+1} - \btheta_t &= -\eta \mathbb{E}_{\mathbf{x} \sim \mathcal{D}}[\nabla_{\btheta_t} \mathcal{L}(\phi_{\btheta_t}(\mathbf{x}))] \\
&= -\eta \mathbb{E}_{\mathbf{x} \sim \mathcal{D}}\left[\nabla_{\btheta_t} \phi_{\btheta_t}(\mathbf{x}) \mathcal{L}'(\phi_{\btheta_t}(\mathbf{x}))\right]
\end{align}

where $\mathcal{L}(\phi_{\btheta_t}(\mathbf{x}))$ represents the loss function and $\eta$ is the learning rate.

The change in the output of the network at step $t$ can be expressed as:
\small
\begin{align}
\nabla \btheta(\mathbf{x}')  &=  \phi_{\btheta_{t+1}}(\mathbf{x}') - \phi_{\btheta_t}(\mathbf{x}') \\   &= \left\langle \nabla_{\btheta_t} \phi_{\btheta_t}(\mathbf{x}'), \btheta_{t+1} - \btheta_t \right\rangle \\
&= -\eta \nabla_{\btheta_t} \phi_{\btheta_t}(\mathbf{x}')^{\top} \mathbb{E}_{\mathbf{x}} \left[ \nabla_{\btheta_t} \phi_{\btheta_t}(\mathbf{x}) \mathcal{L}'(\phi_{\btheta_t}(\mathbf{x})) \right] \\
&= -\eta \mathbb{E}_{\mathbf{x}} \left[\nabla_{\btheta_t} \phi_{\btheta_t}(\mathbf{x}')^{\top} \nabla_{\btheta_t} \phi_{\btheta_t}(\mathbf{x}) \mathcal{L}'(\phi_{\btheta_t}(\mathbf{x})) \right] \\
&= -\eta \mathbb{E}_{\mathbf{x}} \left[\mathbf{K}(\mathbf{x}, \mathbf{x}') \mathcal{L}'(\phi_{\btheta_t}(\mathbf{x})) \right]
\end{align}
\normalsize
where $\mathbf{K}(\mathbf{x}, \mathbf{x}') = \nabla_{\btheta_t} \phi_{\btheta_t}(\mathbf{x})^{\top} \nabla_{\btheta_t} \phi_{\btheta_t}(\mathbf{x}')$ is the NTK matrix at time $t$.

We now proceed to prove by induction that the NTK of the Propulsion method closely approximates the NTK of full fine-tuning. In theory, we introduce a diagonal matrix \( \mathbf{Z} \), and we can write the Propulsion model as:
\[
\phi_P(\mathbf{x}; \btheta) = \btheta_0 \mathbf{x} \odot \mathbf{z} = \btheta_0 \mathbf{x} \mathbf{Z},
\]
where \( \mathbf{Z} \) is a diagonal matrix, and the diagonal elements of \( \mathbf{Z} \) correspond to the Propulsion parameters \( \mathbf{z} \).

\textbf{Base Case:}
Consider the model before training at $t = t_0$. The output of the Propulsion model can be written as:
\[
\phi_P(\mathbf{x}; \btheta_{t_0}) = \btheta_0 \mathbf{x} \mathbf{Z}_0,
\]
where $\btheta_0$ is the pre-trained weight matrix and $\mathbf{Z}_0 = I_n$ is the identity matrix (i.e., initially, the diagonal matrix $\mathbf{Z}$ is an identity matrix). In this case, the gradient with respect to the parameters is:
\[
\nabla \phi_P(\mathbf{x}; \btheta_{t_0}) = \btheta_0 \mathbf{x}.
\]
Since $\mathbf{Z}_0 = I_n$, the gradient is identical to the gradient of the fully fine-tuned model:
\[
\nabla \phi_P(\mathbf{x}; \btheta_{t_0}) = \nabla \phi_F(\mathbf{x}; \btheta_{t_0}).
\]
Thus, the NTK for Propulsion at initialization is identical to the NTK for full fine-tuning:
\[
\mathbf{K}_P(\mathbf{x}, \mathbf{x}') = \mathbf{K}_F(\mathbf{x}, \mathbf{x}').
\]
\textbf{Inductive Hypothesis:}
Assume that at step $t$, the Propulsion model is of the form:
\[
\phi_P(\mathbf{x}; \btheta_t) = \btheta_0 \mathbf{x} \mathbf{Z}_t,
\]
where $\mathbf{Z}_t$ is the updated diagonal matrix at time $t$. The gradient with respect to the diagonal parameters is:
\[
    \nabla \phi_{P}(\mbx ; \btheta_{t}) =  \nabla \phi_{P}(\btheta_0\mbx ; \mathbf{Z}_{t}).
\]

We now compute the NTK for Propulsion at step $t$:
\small
\begin{align}
   \nabla \phi_{P}\left(\mathbf{x}_i;\btheta_{t}\right) \cdot \nabla \phi_{P}(\mathbf{x}_j; \btheta_{t})^{\top} 
   &=   \nabla \phi_{P}(\btheta_0 \mathbf{x}_i ; \btheta_{z_t}) \cdot   \nabla \phi_{P}(\btheta_0 \mathbf{x}_j ; \btheta_{z_t})^{\top} 
\end{align}
\normalsize

Now from the definition of NTK, we can write:
\begin{align}
   \nabla \phi_{P}\left(\mbx_i;\btheta_{t}\right) \cdot \nabla \phi_{P}(\mbx_j; \btheta_{t})^{\top} =  \mbK^P\left(\btheta_0\mbx_i, \btheta_0\mbx_j\right)
\end{align}

\textbf{Inductive Step:}
We now show that the NTK for Propulsion converges to the NTK for full fine-tuning.

From the definition of kernel behavior in the NTK regime, we know that for large $d$, the width of the network, the change in the NTK over time is small. Specifically, for large $d$, we have:
\small
\[
\nabla \phi_P\left(\mathbf{x}; \btheta_t\right) - \nabla \phi_P\left(\mathbf{x}; \btheta_{t_0}\right) \approx \nabla \phi_F\left(\mathbf{x}; \btheta_t\right) - \nabla \phi_F\left(\mathbf{x}; \btheta_{t_0}\right).
\]
\normalsize
Since at $t_0$, we have $\nabla \phi_P(\mathbf{x}; \btheta_{t_0}) = \nabla \phi_F(\mathbf{x}; \btheta_{t_0})$, it follows that:
\begin{align}
\nabla \phi_P(\mathbf{x}; \btheta_t) \approx \nabla \phi_F(\mathbf{x}; \btheta_t).
\end{align}

Thus we can write 
\begin{align}
   \nabla \phi_{F}\left(\mbx_i;\btheta_{t}\right) \cdot \nabla \phi_{F}(\mbx_j; \btheta_{t})^{\top}  \approx \mbK^P\left(\btheta_0\mbx_i, \btheta_0\mbx_j\right)
\end{align}

Which is simply implies 
\begin{align}
   \mbK^F\left(\mbx, \mbx^{\prime}\right)  \approx \mbK^P\left(\btheta_0\mbx_i, \btheta_0\mbx_j\right)
\end{align}

Thus, the NTK for Propulsion approximates the NTK for full fine-tuning:
\[
   \mbK^F\left(\mbx, \mbx^{\prime}\right)  \approx \mbK^P\left(\btheta_0\mbx_i, \btheta_0\mbx_j\right)
\]

\textbf{Error Bound:}
To formalize the error between the NTK for Propulsion and full fine-tuning, we apply the Johnson-Lindenstrauss Lemma. 

Given vectors $\mathbf{u}, \mathbf{v} \in \mathbb{R}^d$ with $\|\mathbf{u}\|, \|\mathbf{v}\| \leq c$, and a random matrix $A \in \mathbb{R}^{d \times k}$ with i.i.d. entries, the lemma states:
\small
\[
\Pr\left[\left| (A \mathbf{u})^\top (A \mathbf{v}) - \mathbf{u}^\top \mathbf{v} \right| \geq c \epsilon \right] \leq 4 \exp\left(- \frac{(\epsilon^2 - \epsilon^3) d}{4}\right).
\]
\normalsize
Using the Johnson-Lindenstrauss lemma, with a probability of at least \( 1 - 4 \exp\left(- \frac{(\epsilon^2 - \epsilon^3) d}{4}\right) \)
Applying this to our NTK matrices, we get:
\small
\[
\Pr\left[\left| (\btheta_0\mbx_i)^\top (\btheta_0\mbx_j) - \mathbf{x_i}^\top \mathbf{x_j} \right|  \right] \geq 1 - 4 \exp\left(- \frac{(\epsilon^2 - \epsilon^3) d}{4}\right)
\]

\normalsize

\begin{figure*}[htp!]
    \centering
    \begin{subfigure}[b]{\textwidth}
        \includegraphics[width=0.95\textwidth]{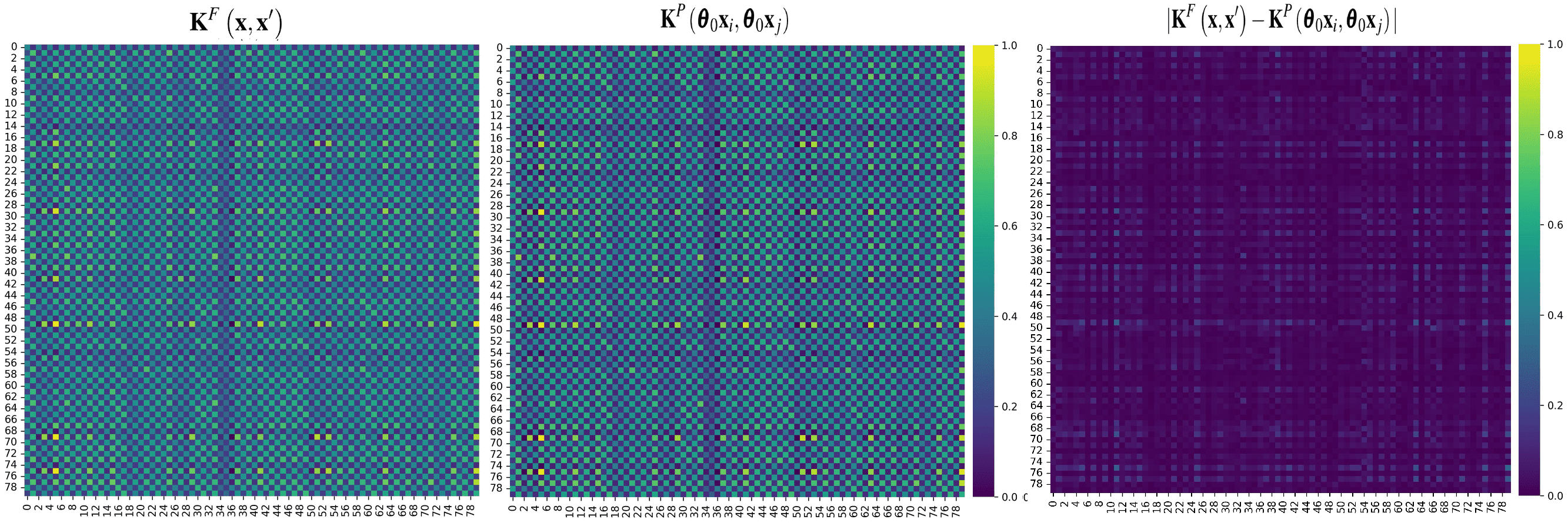}
        \caption{SST-2}
        \label{fig:ntk_sst2}
    \end{subfigure}

    \begin{subfigure}[b]{\textwidth}
        \includegraphics[width=0.95\textwidth]{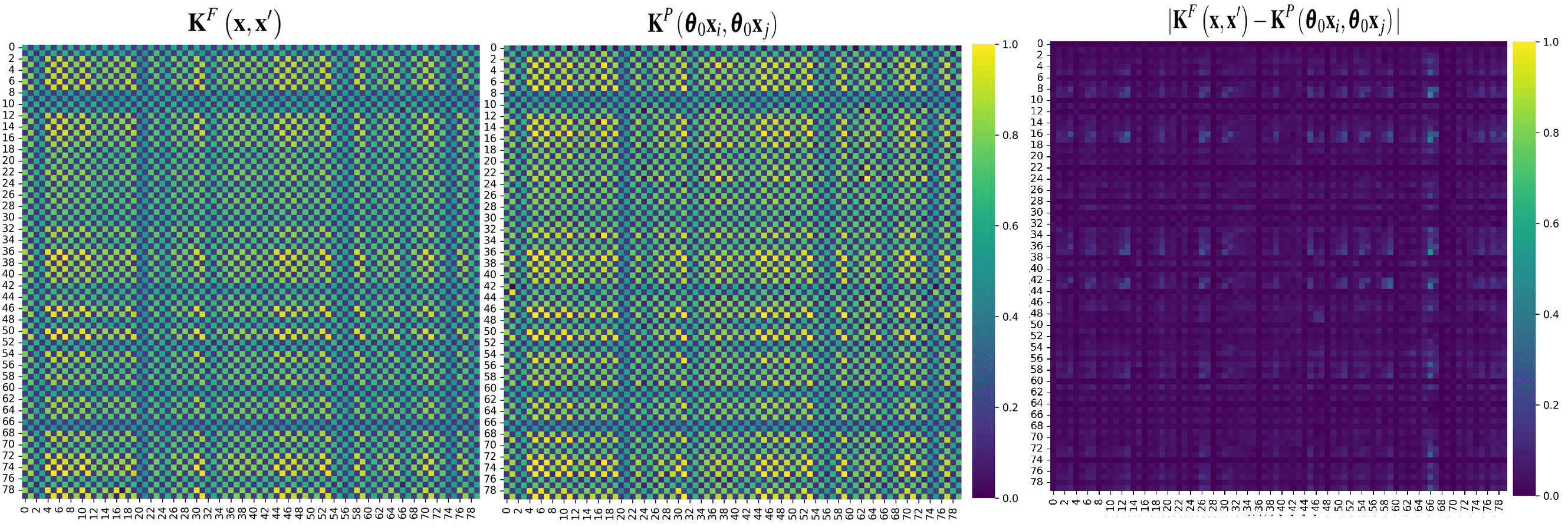}
        \caption{RTE}
        \label{fig:ntk_rte}
    \end{subfigure}
    \begin{subfigure}[b]{\textwidth}
        \includegraphics[width=0.95\textwidth]{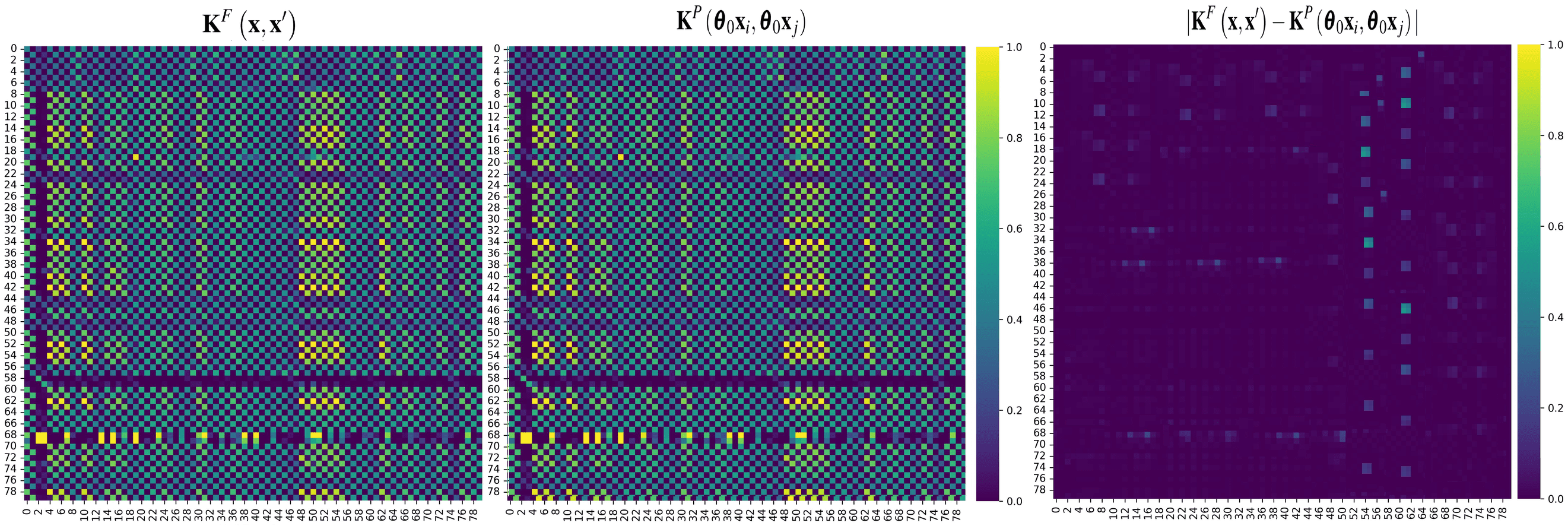}
        \caption{CoLA}
        \label{fig:ntk_cola}
    \end{subfigure}
    \begin{subfigure}[b]{\textwidth}
        \includegraphics[width=0.95\textwidth]{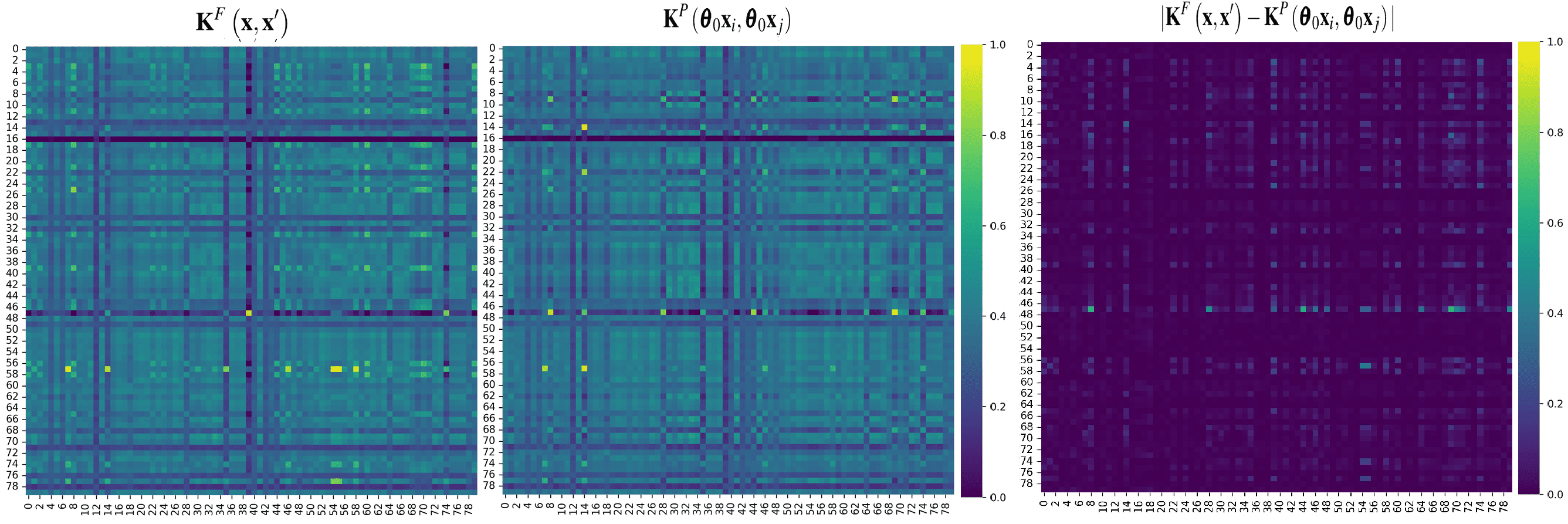}
        \caption{STSB}
        \label{fig:ntk_stsb}
    \end{subfigure}
    \caption{Heat map of NTK matrix on the SST-2, RTE, CoLA, and STSB datasets. For every dataset, the first NTK matrix is from full fine-tuning. The second NTK matrix is from the Propulsion method. The third matrix shows the absolute distance between them. }
    \label{fig:ntk_heatmap}
\end{figure*}

\section{Empirical Validation of NTK Approximation} \label{sec:NTK_results}

In this section, we present empirical evidence to support the theoretical claims made in Theorem \ref{thm:propulsion_ntk}. We compare the NTK matrices of full fine-tuning and Propulsion fine-tuning across four different datasets: SST-2, RTE, CoLA, and STSB. The results, visualized in Figure \ref{fig:ntk_heatmap}, show that the NTK for Propulsion approximates the NTK for full fine-tuning with high accuracy across all datasets.

For each dataset, we compute the NTK matrices using both full fine-tuning and Propulsion fine-tuning. Specifically, the first NTK matrix, denoted as \( \mathbf{K}^F(\mathbf{x}, \mathbf{x}') \), corresponds to the NTK computed from fully fine-tuned models. The second NTK matrix, denoted \( \mathbf{K}^P(\btheta_0 \mathbf{x}, \btheta_0 \mathbf{x}') \), corresponds to the NTK obtained from the Propulsion method, where the base matrix \( \btheta_0 \) remains frozen, and only the task-specific diagonal matrix \( \mathbf{Z} \)  is updated. Finally, to quantify the difference between these two NTK matrices, we compute the absolute distance between them, denoted as \( |\mathbf{K}^F(\mathbf{x}, \mathbf{x}') - \mathbf{K}^P(\btheta_0 \mathbf{x}, \btheta_0 \mathbf{x}')| \). This measures how closely the NTK of Propulsion approximates the NTK of full fine-tuning.

Figure \ref{fig:ntk_heatmap} presents the heatmaps of the NTK matrices across the SST-2, RTE, CoLA, and STSB datasets. The heatmaps are organized as follows: The first column corresponds to \( \mathbf{K}^F(\mathbf{x}, \mathbf{x}') \), the NTK matrix computed from full fine-tuning. The second column corresponds to \( \mathbf{K}^P(\btheta_0 \mathbf{x}, \btheta_0 \mathbf{x}') \), the NTK matrix computed from Propulsion fine-tuning. The third column shows \( |\mathbf{K}^F(\mathbf{x}, \mathbf{x}') - \mathbf{K}^P(\btheta_0 \mathbf{x}, \btheta_0 \mathbf{x}')| \), the absolute difference between the two NTK matrices. The heatmaps demonstrate that the NTK matrices for Propulsion fine-tuning closely resemble those for full fine-tuning across all four datasets. The third column, which shows the absolute difference, indicates that the discrepancies between the two methods are minimal. This supports our theoretical findings in Section \ref{thm:propulsion_ntk}, which claim that Propulsion approximates full fine-tuning under the NTK regime with high probability.

These empirical results validate the claim that Propulsion, despite updating only a diagonal matrix \( \mathbf{Z} \), can closely approximate the behavior of full fine-tuning in the NTK regime. This is particularly significant given that Propulsion fine-tunes a far smaller number of parameters than full fine-tuning, leading to more efficient training while maintaining comparable performance. The minimal difference observed in the third column of Figure \ref{fig:ntk_heatmap} confirms that the theoretical bound on the NTK difference, as stated in Theorem \ref{thm:propulsion_ntk}, holds in practice.

\begin{figure*}[htp!]
    \centering
    \begin{subfigure}[b]{\textwidth}
        \includegraphics[width=0.95\textwidth]{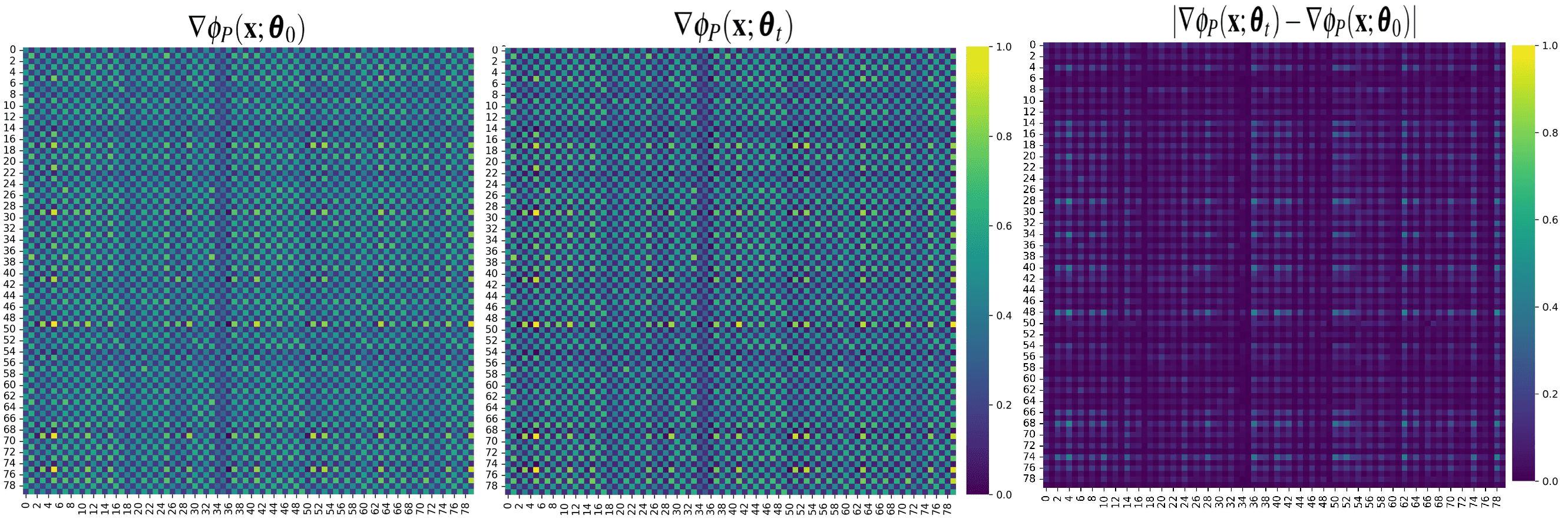}
        \caption{SST-2}
        \label{fig:k_sst2}
    \end{subfigure}

    \begin{subfigure}[b]{\textwidth}
        \includegraphics[width=0.95\textwidth]{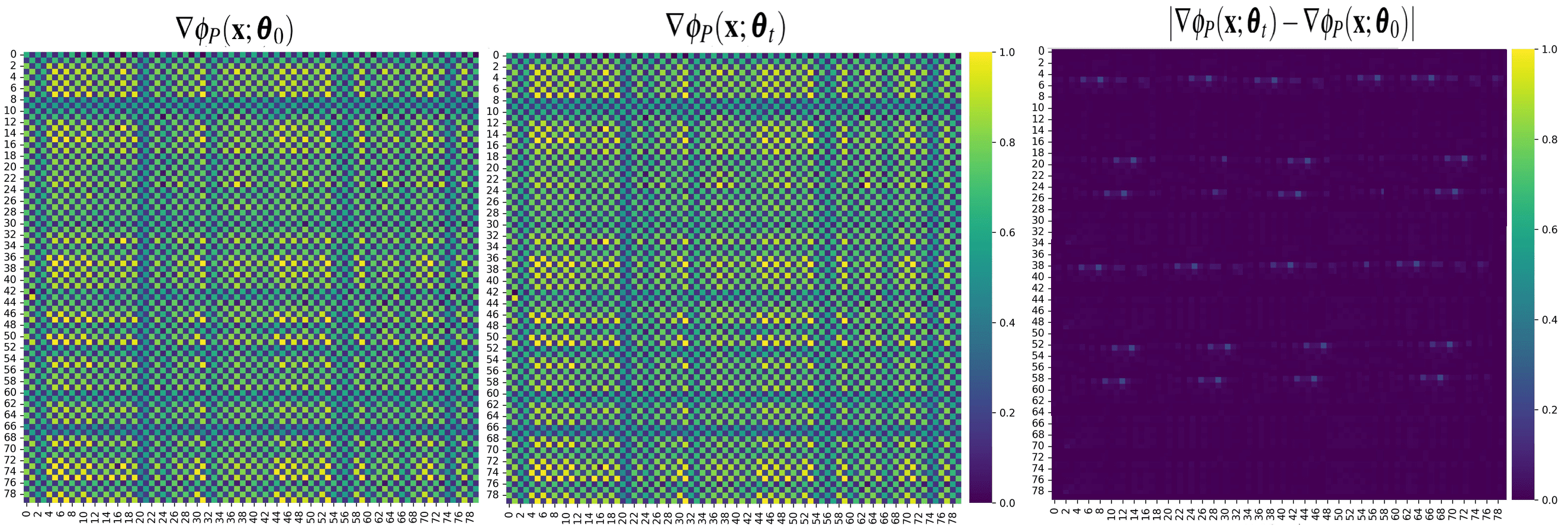}
        \caption{RTE}
        \label{fig:k_rte}
    \end{subfigure}
    \begin{subfigure}[b]{\textwidth}
        \includegraphics[width=0.95\textwidth]{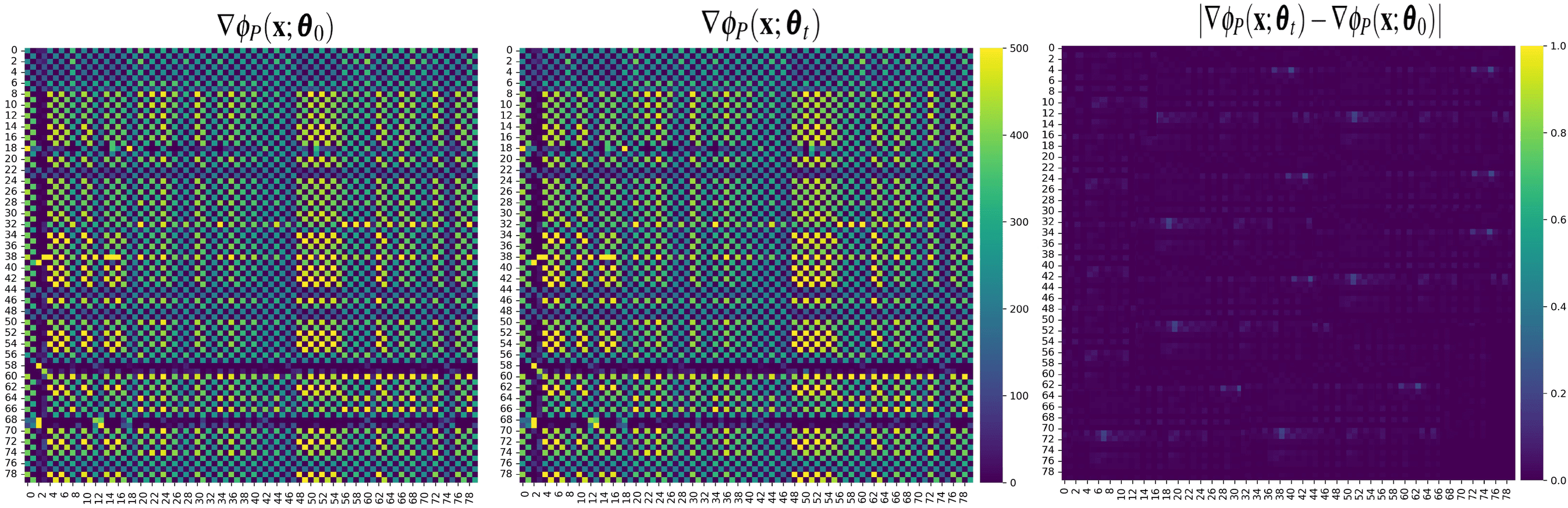}
        \caption{CoLA}
        \label{fig:k_cola}
    \end{subfigure}
    \begin{subfigure}[b]{\textwidth}
        \includegraphics[width=0.95\textwidth]{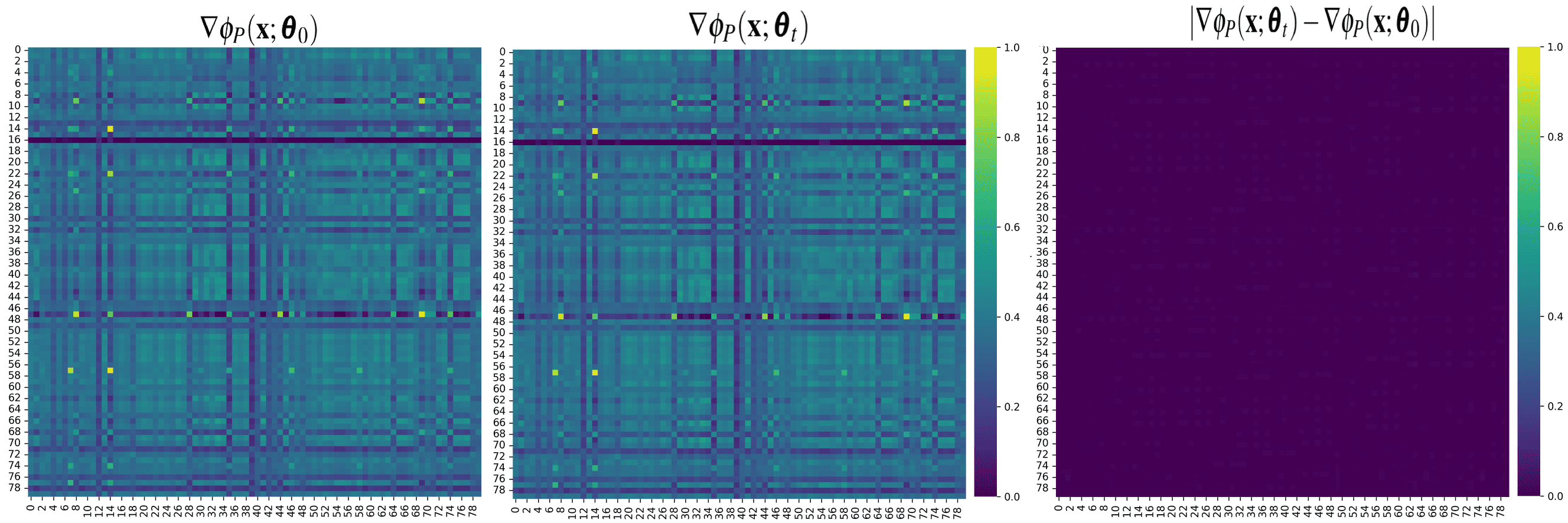}
        \caption{STSB}
        \label{fig:k_stsb}
    \end{subfigure}
    \caption{Heat map of Jacobian matrix on the SST-2, RTE, CoLA, and STSB datasets. For every dataset, the first Jacobian matrix is from the initial steps before training. The second Jacobian matrix is from the \( t \)-steps of training. The third matrix shows the absolute distance between them. }
    \label{fig:jacobian_heatmap}
\end{figure*}

\section{Kernel Behavior in the NTK Regime} \label{sec:kernel_behaviour}

In this section, we provide empirical validation of the kernel behavior in the NTK regime. As the width of the neural network tends to infinity, the gradient of the network's output with respect to its parameters stabilizes, and the network exhibits linear behavior in the parameter space. This property of NTK is crucial for understanding the training dynamics of neural networks, particularly in fine-tuning scenarios such as Propulsion.

We evaluate the kernel behavior by analyzing the Jacobian matrix of the network's output with respect to the parameters \( \btheta_0 \) before and after several steps of training. Specifically, we compute the gradient of the model output \( \phi_{\btheta}(\mathbf{x}) \) with respect to the initial parameters \( \btheta_0 \), and compare it to the gradient after \( t \) steps of training, denoted by \( \btheta_t \). For each dataset, the Jacobian matrices are computed as \( \nabla_{\btheta_0} \phi(\mathbf{x}) \) (the initial Jacobian matrix) and \( \nabla_{\btheta_t} \phi(\mathbf{x}) \) (the Jacobian matrix after \( t \) steps of training). To quantify the change in the gradients, we compute the absolute difference between the two Jacobian matrices, \( \left|\nabla_{\btheta_t} \phi(\mathbf{x}) - \nabla_{\btheta_0} \phi(\mathbf{x})\right| \), which measures the stability of the gradients in the NTK regime and indicates whether the network remains in the kernel regime as training progresses.

Figure \ref{fig:jacobian_heatmap} presents the heatmaps of the Jacobian matrices across the SST-2, RTE, CoLA, and STSB datasets. Each row corresponds to one of the datasets and includes three columns: the first column shows \( \nabla_{\btheta_0} \phi(\mathbf{x}) \), the Jacobian matrix computed from the initial model parameters before training. The second column shows \( \nabla_{\btheta_t} \phi(\mathbf{x}) \), the Jacobian matrix computed after \( t \) steps of training. The third column shows the absolute difference between the two Jacobian matrices, \( \left|\nabla_{\btheta_t} \phi(\mathbf{x}) - \nabla_{\btheta_0} \phi(\mathbf{x})\right| \). The heatmaps demonstrate that the Jacobian matrices remain relatively stable after \( t \) steps of training across all datasets. This suggests that the gradients are largely unchanged, confirming that the network is operating in the NTK regime, where the parameters exhibit kernel behavior, and the network's output becomes a linear function of the parameters.

The kernel behavior observed in the Jacobian matrices across different datasets aligns with the theoretical understanding of the NTK regime. In this regime, the network's output becomes a function of the NTK matrix, and the gradients with respect to the parameters stabilize as the width of the network increases. The results in Figure \ref{fig:jacobian_heatmap} provide empirical evidence that, even after several steps of training, the gradients remain close to their initial values, indicating that the network has not deviated significantly from the kernel regime. This behavior is particularly relevant for fine-tuning methods like Propulsion, where the stability of gradients ensures that the model can be fine-tuned efficiently without large deviations from the pre-trained parameters. The minimal differences observed in the third column of the heatmaps confirm that the kernel behavior holds in practice, and the network remains in the NTK regime as training progresses.

\begin{table*}[htp!]
\centering
\scalebox{0.75}{
\begin{tabular}{l|l|p{10cm}}
\hline
\textbf{Method}        & \textbf{$\Delta W$ Reparameterization} & \textbf{Notes} \\ \hline
Intrinsic SAID         & $\Delta W  = F(W^r)$   & F : $\mathbb{R}^r \rightarrow \mathbb{R}^d$, $W^r \in \mathbb{R}^r$ are parameters to be optimized, and $r \ll d$. \\ \hline
LoRA                   & $\Delta W = W_{\text{down}} W_{\text{up}}$  & $W_{\text{down}} \in \mathbb{R}^{d \times r}$, $W_{\text{up}} \in \mathbb{R}^{r \times d}$, and $r \ll \{k,d\}$. \\ \hline
KronA                  & $\Delta W = W_{\text{down}} \otimes W_{\text{up}}$ & rank($W_{\text{down}} \otimes W_{\text{up}}$) = rank($W_{\text{down}}$) $\times$ rank($W_{\text{up}}$). \\ \hline
DyLoRA                 & $\Delta W = W_{\text{down}\downarrow b} W_{\text{up}\downarrow b}$ & $W_{\text{down}\downarrow b} = W_{\text{down}}[:b,:], \quad W_{\text{up}\downarrow b} = W_{\text{up}}[:, :b], \quad b \in \{r_{\text{min}}, \dots, r_{\text{max}}\}$. \\ \hline
AdaLoRA                & $\Delta W  = PAQ$ & $PP^\top = P^\top P \neq I = QQ^\top = Q^\top Q$, $\Lambda = \text{diag}(\sigma_1, \sigma_2, \dots, \sigma_r)$. \\ \hline
IncreLoRA              & $\Delta W = W_{\text{down}} \Lambda W_{\text{up}}$ & $\Lambda = [\lambda_1, \lambda_2, \dots, \lambda_r]$ with $\lambda_i$ being an arbitrary constant. \\ \hline
DeltaLoRA              & $\Delta W = W_{\text{down}} W_{\text{up}}$ & $W^{(t+1)} \gets W^{(t)} + \big(W_{\text{down}}^{(t+1)} W_{\text{up}}^{(t+1)} - W_{\text{down}}^{(t)} W_{\text{up}}^{(t)}\big)$. \\ \hline
LoRAPrune              & $\Delta W = W_{\text{down}} W_{\text{up}} \odot M$ & $\delta = (W + W_{\text{down}} W_{\text{up}}) \odot M, \quad M \in \{0, 1\}^{1 \times G}, \quad G \text{ is group number.}$ \\ \hline
QLoRA                  & $\Delta W = W_{\text{down}}^{BF16} W_{\text{up}}^{BF16}$ & $Y^{BF16} = X^{BF16} \cdot \text{doubleDequant}(c_1^{FP32}, c_2^{FP8}, W^{NF4}) 
+ X^{BF16} W_{\text{down}}^{BF16} W_{\text{up}}^{BF16}$. \\ \hline
QA-LoRA                & $\Delta W = W_{\text{down}} W_{\text{up}}$ & $W_{\text{down}} \in \mathbb{R}^{d \times r}$, $W_{\text{up}} \in \mathbb{R}^{r \times L}$, $L$ is the quantization group number of W. \\ \hline
LoFTQ                  & $\Delta W = SVD(W - Q_t$) & $Q_t = q_N \big(W - W_{\text{down}}^{t-1} W_{\text{up}}^{t-1} \big), \quad q_N \text{ is } N\text{-bit quantization function.}$ \\ \hline
Kernel-mix             & $\Delta W^h = \big(B_{\text{LoRA}}, B^h\big) 
\begin{pmatrix}
A_{\text{LoRA}}^h \\ 
A^h
\end{pmatrix}$ & $B_{\text{LoRA}} \text{ is shared across all heads, } B^h, A^h \text{ provide rank-}r  \text{ update  in each head.}$. \\ \hline
LoRA-FA                & $\Delta W = W_{\text{down}} W_{\text{up}} = Q R W_{\text{up}}$ & $W_{\text{down}}$ is frozen, and only $W_{\text{up}}$ is updated. \\ \hline
Propulsion               & $\Delta W = W\odot Z$ & $W$ is frozen, and only $Z$ is updated. \\ \hline
\end{tabular}
}
\caption{Comparison of delta weight reparameterization across various PEFT methods. Representations of the baseline methods are taken from \citet{xu2023parameter}.} \label{tab:delta_cmp}
\end{table*}

\section{Comparison of Delta Weight Reparameterization in PEFT Methods} \label{sec:delta_cmp}
Table \ref{tab:delta_cmp} provides a comprehensive comparison of various PEFT methods based on their reparameterization of the delta weight matrix \( \Delta W \). Each method uses different strategies for adjusting the weight updates during fine-tuning, optimizing parameter efficiency while maintaining performance. 

For example, methods like LoRA and KronA employ low-rank decompositions, while methods like Propulsion, introduced in this work, use element-wise updates, where the base weights \( W \) remain frozen and only the task-specific matrix \( Z \) is updated. This comparison highlights the diverse approaches used across methods, showing how the trade-off between memory efficiency and computational complexity is handled.

\section{Training} \label{sec:training}

Algorithm \ref{algo:orpulsion} describes the training process of the \emph{Propulsion} method. We begin with an input \( x \) and a pre-trained language model \( \mathbb{M}(.) \) consisting of \( L \) layers, where all parameters of \( \mathbb{M}(.) \) are frozen. The \emph{Propulsion} parameters \( \mathcal{Z} \) are initialized at the beginning of training. During each training epoch, the output \(V\) is extracted from a given layer \(L_{i}\). Output \(V\) is then updated to \(V^{\prime}\) through element-wise multiplication with \(\mathbf{z_i}^k\).  This new transformed output of a given layer \(L_{i}\) is then sent through the rest of the model, where it is used as the input \(x\) for the subsequent layer \(L_{i+1}\), where \(i\) ranges from \(1\) to \( N \). After processing the input through all layers, the loss specific to the task is calculated, and the \emph{Propulsion} parameters \( \mathcal{Z}\) are updated based on this loss.

% \Jia{remove the repetition for example explaining the push and others.}\Hongbo{Addressed}
% \Jia{make the description more clear. give examples.}

\begin{algorithm}[!t]
\caption{Propulsion PEFT training}\label{algo:orpulsion}
\begin{algorithmic}
\Require  input $x$, a retrained LM model  $\mathbb{M}(.)$ with $L$ layers
\Ensure Freeze all parameters of $\mathbb{M}(.)$
\Ensure Initialization Propulsion parameters $\mathcal{Z}$

\While{$epoch < epochs$}
\For{$i \gets 1$ to $N$}
    \State $V = L_{i}(x)$ \Comment{Output of layer $L_i$}
    
     \State{$  V^\prime = [\mathbf{v_j} \odot \mathbf{z}_i^k]_{j=1}^{s}$}\Comment{Updating output}
     \State{$ x \gets  V^\prime$}
  
\EndFor

    \State $\text{Calculating loss for task specific goal}$
    \State $\text{update parameters}~ \mathcal{Z}$

\EndWhile
\end{algorithmic}
\end{algorithm}

After we employ the \emph{Propulsion} method to modify the outputs at all layers and fine-tune the model, we calculate the loss. We update only the \emph{Propulsion} parameters $\mathcal{Z}$, based on the task-specific loss - the other parameters within the model remain frozen. For STS-B dataset, we have used Mean Squared Error and rest of all experiments in this study, we utilize cross-entropy loss as our objective function, which is defined below:
\begin{equation}
\mathcal{L}(\mathbf{y}, \mathbf{\hat{y}}) = -\frac{1}{T} \sum_{t=1}^{T} \mathbf{y}_t \log(\mathbf{\hat{y}}_t)
\label{eq:cross_entropy_loss}
\end{equation}
where, $T$ represents the total number of data samples, $\mathbf{y}$ is the ground truth, and $\mathbf{\hat{y}}$ are the predicted labels.  Although we focused on Transformer-based pre-trained language models to test the \emph{Propulsion} method, it can be applied to any pre-trained Neural Network for PEFT fine-tuning because it modifies the output of each layer, independent of the model structure.

\section{More Ablation Study} \label{sec:moreabls}
\begin{figure}[!t]
%\begin{wrapfigure}{r}{0.7\textwidth}
%\captionsetup{font=footnotesize}

 \begin{center}

      \includegraphics[width=0.95\linewidth]{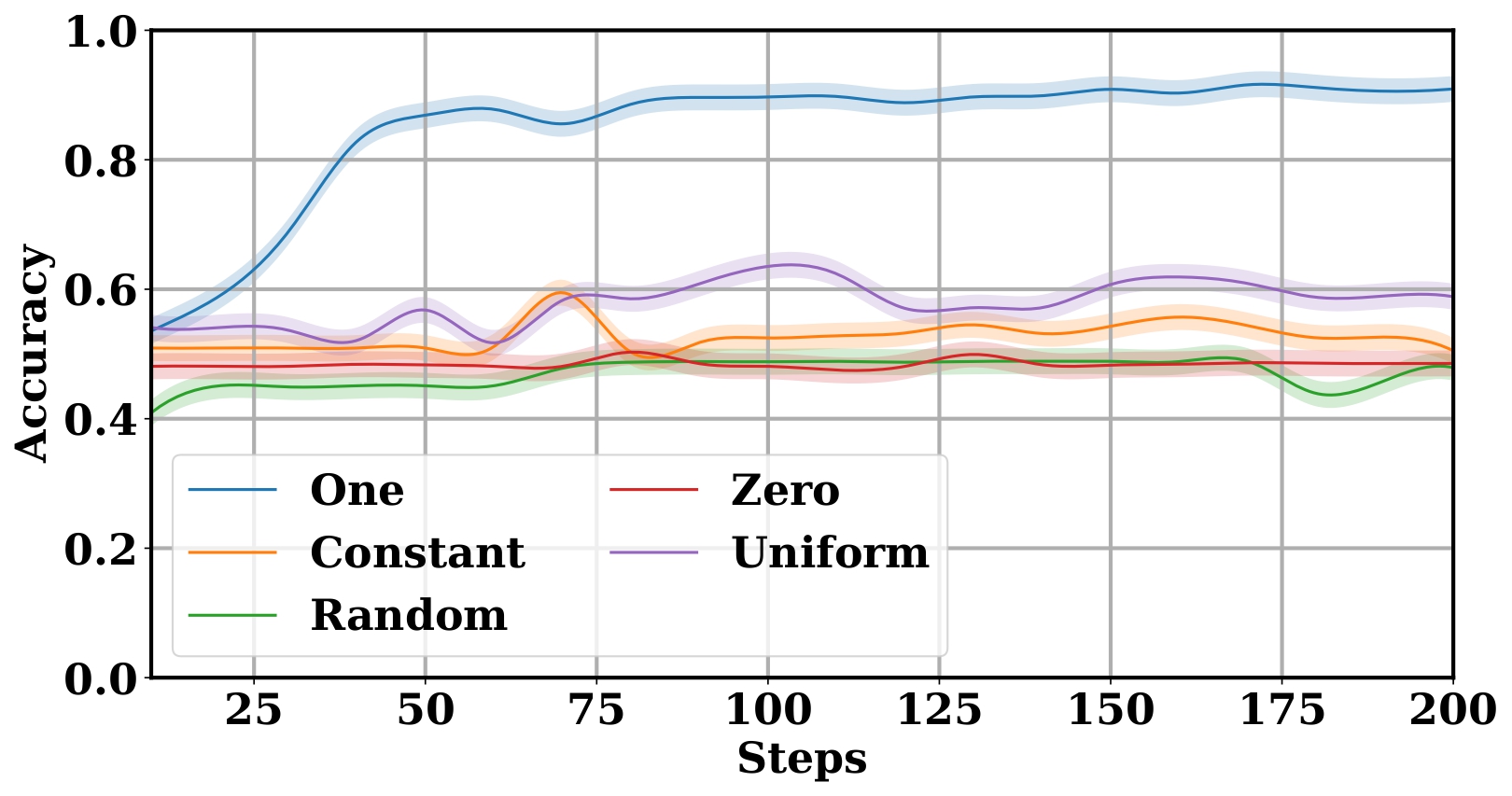}
   \end{center}
%\end{wrapfigure}
\caption{Performance comparison of Propulsion parameter initialization techniques}
\label{fig:initilization}
\end{figure}
\begin{figure*}[!t]
%\begin{wrapfigure}{r}{0.7\textwidth}
%\captionsetup{font=footnotesize}
   
 \begin{center}

      \includegraphics[width=0.85\linewidth]{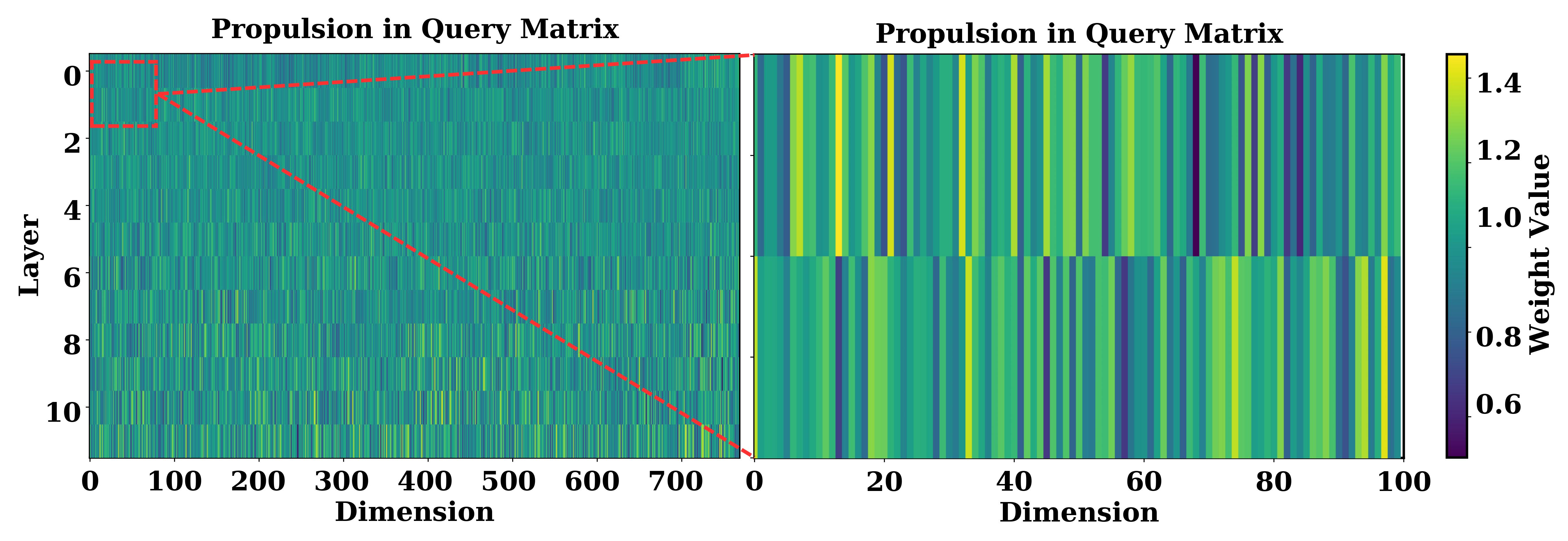}
   \end{center}

%\end{wrapfigure}
\caption{Visualization of trained \emph{Propulsion} parameters across the attention query layers after fine-tuning on the SST-2 dataset. Each layer and dimension is represented, indicating the diversity of weight adjustments necessary for task-specific performance optimization.}
\label{fig:Propulsion_score}
\end{figure*}

\textbf{Propulsion Parameter Initialization:} Setting \emph{Propulsion} parameters correctly is important for the model to operate accurately and efficiently. As shown in Figure \ref{fig:initilization}, we tested different methods to set these parameters on the SST-2 dataset. The results clearly show that initializing the \emph{Propulsion} parameter to $1$ gives the best performance. This superior performance can be explained by the behavior of the model during the first forward pass. Specifically, when the \emph{Propulsion} parameter is set to $1$, it ensures that the output of each layer in the initial forward pass remains identical to that without any \emph{Propulsion} modification. This approach allows the model to operate from a well-understood and predictable starting point. It uses the original output projection, which is a familiar projection of the behavior of the model, thereby facilitating smoother subsequent updates and adjustments to the \emph{Propulsion} parameters.

\textbf{Propulsion Weights After Training: } In Figure \ref{fig:Propulsion_score},  we observe that the \emph{Propulsion} parameter weightings across different dimensions and layers are a crucial aspect of our analysis. Initially, the \emph{Propulsion} weights are set to $1$, and after training, they range between $0.98$ and $1.02$. This variation suggests that a small adjustment to the projection of the layer output is necessary to achieve a task-specific goal. The left side of the figure depicts the distribution of the \emph{Propulsion} weights across all dimensions and layers at the start of the training, which shows uniformly set weights of $1$. The right side of the figure, which focuses on a subset of dimensions, illustrates the distribution of \emph{Propulsion} weights after training, displaying the variation in the weights. This variation indicates that the model fine-tunes the \emph{Propulsion} parameter to optimize performance, reflecting the specific requirements of the task. These observations highlight the significance of allowing small adjustments to the \emph{Propulsion} parameter. Even minor changes in weight can significantly impact the model's ability to meet task-specific goals. Hence, the \emph{Propulsion} parameter plays an important role in the fine-tuning process and contributes to the overall performance of the model.

\subsection{Multi-Propulsion}

Instead of utilizing a single \emph{Propulsion}  vector in a layer, we can employ multiple \emph{Propulsion}  vectors to gain more control over the model’s adjustments by following a pooling operation. This pooling operation dynamically synthesizes the influence of these vectors, effectively combining their effects into a single output matrix \(V_i'\). If we use the total $p$ numbers of \emph{Propulsion}, then we can define the pooling operation as:
\[
V_i' = \text{Pooling}(V_i^{1'}, V_i^{2'}, \ldots, V_i^{p'})
\]
The pooled output \(V_i'\) is then processed as the input for the subsequent layer \(L_{i+1}\), or can be adjusted according to specific model requirements or task-based needs.

\begin{figure}[!t]
%\begin{wrapfigure}{r}{0.7\textwidth}
%\captionsetup{font=footnotesize}

 \begin{center}

      \includegraphics[width=0.95\linewidth]{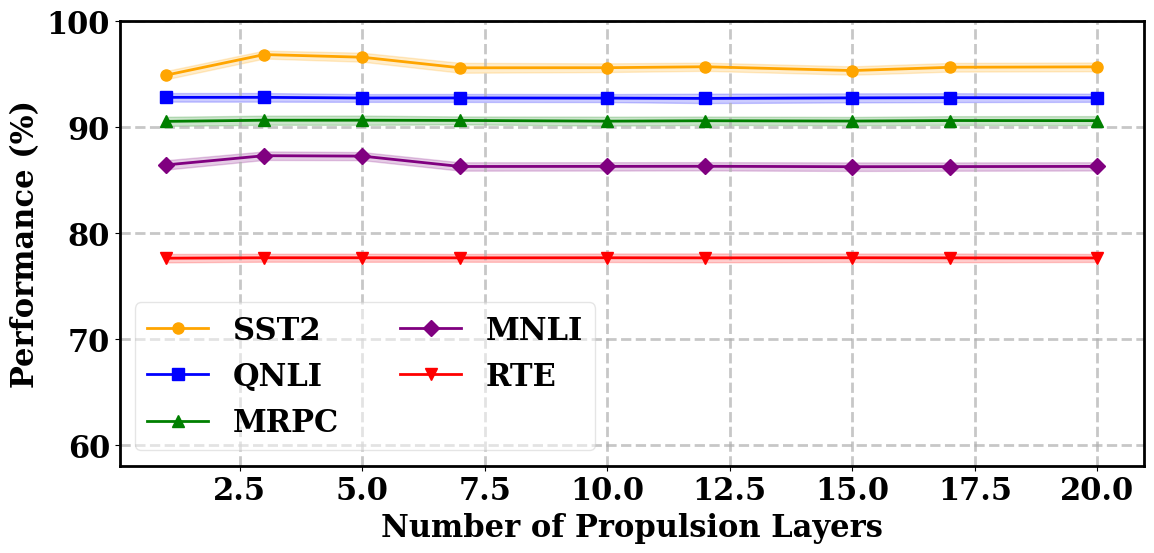}
   \end{center}
%\end{wrapfigure}
\caption{Model performance across five NLP benchmarks (SST2, QNLI, MRPC, MNLI, RTE) with SST2 at ~95\% accuracy and RTE steady at ~80\% across \emph{Propulsion} units (1 to 20.0)}
\label{fig:multi_prop}
\end{figure}

\textbf{Number of Propulsion Layers: } We evaluate our model's performance on five prominent NLP benchmarks: \textit{SST2}, \textit{QNLI}, \textit{MRPC}, \textit{MNLI}, and \textit{RTE}. As shown in Figure~\ref{fig:multi_prop}, our model maintains high accuracy across varying \emph{Propulsion} layer counts (1 to 20.0). \textit{SST2} achieves the highest accuracy, consistently near 95\%, while \textit{RTE} remains stable at around 80\%. Across datasets, performance does not significantly fluctuate with more Propulsion layers, indicating that this method delivers robust performance across diverse tasks.

\begin{figure}[!t]
%\begin{wrapfigure}{r}{0.7\textwidth}
%\captionsetup{font=footnotesize}

 \begin{center}

      \includegraphics[width=0.93\linewidth]{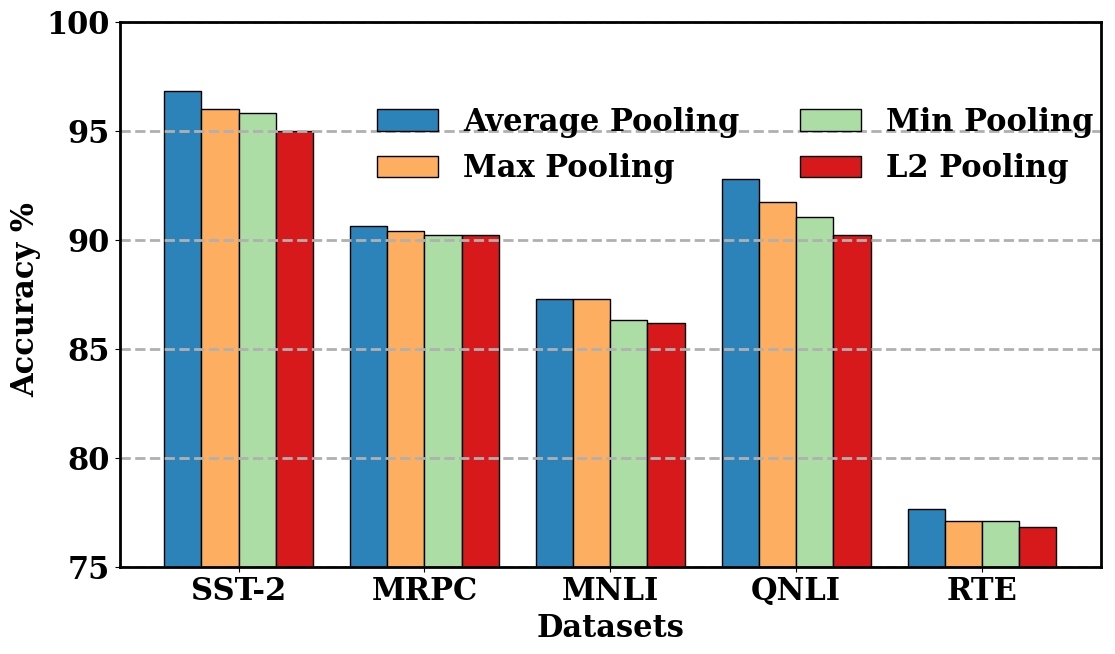}
   \end{center}
%\end{wrapfigure}
\caption{Accuracy comparison of pooling strategies (Average, Max, Min, L2) across five NLP datasets (SST-2, MRPC, MNLI, QNLI, RTE). Average Pooling consistently achieves the highest accuracy, while L2 Pooling tends to underperform.}
\label{fig:pooling_comparison}
\end{figure}

\textbf{Pooling Comparison : } We evaluate the impact of four pooling strategies—\textit{Average}, \textit{Max}, \textit{Min}, and \textit{L2}—on model accuracy across five benchmark datasets: \textit{SST-2}, \textit{MRPC}, \textit{MNLI}, \textit{QNLI}, and \textit{RTE}. Figure \ref{fig:pooling_comparison} compares the different pooling methods across datasets, with \textit{Average Pooling} consistently delivering the highest accuracy, achieving 96.83\% on \textit{SST-2} and 92.79\% on \textit{QNLI}, outperforming \textit{Max}, \textit{Min}, and \textit{L2 Pooling} by up to 1.06\%. On \textit{MRPC} and \textit{MNLI}, all pooling methods perform similarly, though \textit{Average Pooling} maintains a slight edge. In the more challenging \textit{RTE} dataset, differences are minimal, with \textit{Average Pooling} at 77.64\% and \textit{L2 Pooling} at 76.83\%. These results demonstrate that \textit{Average Pooling} provides the best generalization across various text classification tasks.

\section{Baseline Methods}
\textbf{Full Finetuning (FT):} \cite{zhang2022adaptive} Full fine-tuning entails updating all pre-trained weights of a language model with task-specific data. This enables the model to learn intricate patterns, particularly specific tasks, although it requires substantial computational resources and labeled data. However, this process can result in overfitting, particularly when the task-specific dataset is limited or the model is already well suited for the target task.
\textbf{Adapter\textsuperscript{S}:
}\cite{houlsby2019parameter} is a fine-tuning method that involves incorporating task-specific adapter modules into a pretrained model. This approach allows parameter-efficient tuning without requiring extensive modifications to the weights of the original model. These adapters are often characterized by their low-rank properties and include a non-linear activation function that facilitates task-specific adjustments while preserving a significant portion of the pre-trained parameters.

\textbf{Prompt tuning:} \cite{lester2021power} Prompt-tuning entails appending trainable prompt tokens to the input of a language model, thereby updating only the prompt parameters through gradient descent while leaving the pretrained model's parameters frozen, which makes it a memory-efficient approach for fine-tuning. The success of prompt tuning is highly contingent upon the length and training of prompt tokens.

\textbf{Prefix-tuning:} \cite{li2021prefix} Prefix-tuning is an extension of prompt tuning that introduces task-specific vectors into the activations of the multi-head attention layers of the model. These prefixes are optimized independently and do not modify the original pretrained parameters. Prefix-tuning achieves fine-tuning efficiency and stability through a parameterized feed-forward network that parameterizes prefixes.

\textbf{(IA)\textsuperscript{3}:} \cite{liu2022few} The (IA)\textsuperscript{3} approach, which signifies Infused Adapter through Inhibiting and Amplifying Inner Activations, involves element-wise multiplication of model activations with task-specific vectors that have been learned. This strategy facilitates effective adaptation to mixed-task batches without necessitating substantial alterations to the architectural structure of the model, thereby preserving its efficiency and retaining its original form.

\textbf{Bitfit:} \cite{zaken2021bitfit} Bitfit employs a highly parameter-efficient method during fine-tuning, because it selectively updates only the bias parameters of a model. This technique capitalizes on the minimal number of parameters necessary to modify the model outputs, thereby minimizing the memory and computational resources required for full model training.
%\\\textbf{Child-Tuning\textsubscript{ D}:} write here
\textbf{LoRA:} \cite{hu2021lora} Low-Rank Adaptation (LoRA) is a technique that fine-tunes a model by making low-rank updates to the weight matrices, enabling efficient adaptation with minimal alterations to the original parameters. This approach effectively combines the efficiency of parameter utilization and performance in subsequent tasks.

\textbf{AdaLoRA:} \cite{zhang2023adaptive} AdaLoRA is built upon LoRA and enhances its capabilities by adaptively allocating the rank and budget of updates among different weight matrices based on their importance. This approach improves both fine-tuning efficiency and task-specific performance. By dynamically adjusting the rank of the updates and concentrating on the most impactful parameters, AdaLoRA achieves a more effective outcome.

\textbf{MAM Adapter:} \cite{he2021towards} The MAM Adapter integrates the principles of parallel adapter  and prefix-tuning into a cohesive structure. Its objective is to improve model adaptation through optimized parameter allocation, and it is designed to refine various aspects of the model outputs by adjusting a combination of parameters across multiple layers.

\textbf{ProPETL:} \cite{zeng2023one} These techniques are a set of hybrid fine-tuning methods that combine the aspects of adapters, prefix-tuning, and LoRA to optimize the performance across multiple tasks. By integrating multiple strategies into a cohesive approach, these methods aim to leverage the strengths of each technique, while mitigating their weaknesses.

\section{Evaluation Metric}

In this section, we detail the evaluation metrics used to assess the performance of our models across various tasks in the GLUE benchmark suite. Each task is evaluated using specific metrics tailored to its characteristics.

For the CoLA task, we use the Matthews correlation coefficient (MCC) as the evaluation metric. MCC is particularly useful for evaluating binary classification tasks, as it considers into account true and false positives and negatives, providing a balanced measure even with imbalanced datasets.

\[
\text{\scriptsize MCC} = \frac{\text{\scriptsize TP} \times \text{\scriptsize TN} - \text{\scriptsize FP} \times \text{\scriptsize FN}}{\sqrt{(\text{\scriptsize TP} + \text{\scriptsize FP})(\text{\scriptsize TP} + \text{\scriptsize FN})(\text{\scriptsize TN} + \text{\scriptsize FP})(\text{\scriptsize TN} + \text{\scriptsize FN})}}
\]

\vspace{-1em} % Reduce the space after the equation
where:
\begin{itemize}
    \setlength\itemsep{-0.5em} % Reduce space between items
    \item $TP$ = True Positives
    \item $TN$ = True Negatives
    \item $FP$ = False Positives
    \item $FN$ = False Negatives
\end{itemize}

The MRPC and QQP tasks are both designed to assess the ability of a model to determine whether two sentences are semantically equivalent. To evaluate the performance of a model on these tasks, two metrics are used: accuracy and F1 score. Accuracy measures the percentage of correctly identified paraphrase pairs, while the F1 score provides a balance between precision and recall, offering a more nuanced view of the model's performance in identifying paraphrases.

However, the MNLI task requires the model to classify sentence pairs into one of three categories: entailment, contradiction, or neutral. To evaluate the model's performance on this task, the Average Matched Accuracy is reported, which measures the model's accuracy on the matched validation set (in-domain data). This metric reflects the model's ability to generalize across different genres, providing insights into its robustness and versatility.

\subsection*{Accuracy}
\[
\text{\scriptsize Accuracy} = \frac{\text{\scriptsize TP} + \text{\scriptsize TN}}{\text{\scriptsize TP} + \text{\scriptsize TN} + \text{\scriptsize FP} + \text{\scriptsize FN}}
\]

\subsection*{F1 Score}
\[
\text{\scriptsize F1} = 2 \times \frac{\text{\scriptsize Precision} \times \text{\scriptsize Recall}}{\text{\scriptsize Precision} + \text{\scriptsize Recall}}
\]

\vspace{-1em} % Reduce the space after the equations
where:
\begin{itemize}
    \setlength\itemsep{-0.5em} % Reduce space between items
    \item $\text{\scriptsize Precision} = \frac{\text{\scriptsize TP}}{\text{\scriptsize TP} + \text{\scriptsize FP}}$
     \item $\text{\scriptsize Recall} = \frac{\text{\scriptsize TP}}{\text{\scriptsize TP} + \text{\scriptsize FN}}$
\end{itemize}

For the STS-B task, which involves predicting the degree of semantic similarity between sentence pairs, we use both Pearson and Spearman correlation coefficients to evaluate performance. These metrics measure the linear and rank correlations between the predicted and actual similarity scores, respectively.

\subsection*{Pearson Correlation}
\[
r = \frac{\sum (x_i - \bar{x})(y_i - \bar{y})}{\sqrt{\sum (x_i - \bar{x})^2 \sum (y_i - \bar{y})^2}}
\]

\subsection*{Spearman Correlation}
\[
\rho = 1 - \frac{6 \sum d_i^2}{n(n^2 - 1)}
\]
where $d_i$ is the difference between the ranks of corresponding values and $n$ is the number of pairs.

\section{Dataset Description}

The datasets used in this study are listed in Table \ref{tab:DatasetDescription} and \ref{tab:datasets}.

\begin{table}[htp!]
\centering
\begin{tabular}{l|lcc}
\hline
\textbf{Dataset} & \textbf{Domain} & \textbf{ Train} & \textbf{ Test} \\ \hline
MultiArith & Math & -- & 600 \\ 
AddSub & Math & -- & 395 \\ 
GSM8K & Math & 8.8K & 1,319 \\ 
AQuA & Math & 100K & 254 \\ 
SingleEq & Math & -- & 508 \\ 
SVAMP & Math & -- & 1,000 \\ 
BoolQ & CS & 9.4K & 3,270 \\ 
PIQA & CS & 16.1K & 1,830 \\ 
SIQA & CS & 33.4K & 1,954 \\ 
HellaSwag & CS & 39.9K & 10,042 \\
WinoGrande & CS & 63.2K & 1,267 \\ 
ARC-e & CS & 1.1K & 2,376 \\ 
ARC-c & CS & 2.3K & 1,172 \\
OBQA & CS & 5.0K & 500 \\ 
\end{tabular}
\caption{Details of datasets being evaluated. Math: arithmetic reasoning. CS: commonsense reasoning.}
\label{tab:datasets}
\end{table}

\begin{table}[htp!]
    \centering
    \begin{tabular}{c|ccc} % Added another 'r' for the new column
        \toprule
        Dataset &  \textbf{Train} &  \textbf{Validation} &   \textbf{Test} \\ % Added the column header
        \midrule
        SQuAD v1.1 & 87.6k & 10.6k & - \\
        SQuAD v2.0 & 130k & 11.9k & - \\
        XSum & 204k & 11.3k & 11.3k \\
        DailyMail & 287k & 13.4k & 11.5k \\
        CoLA & 8.55k & 1.04k & 1.06k \\
        SST2 & 67.3k & 872 & 1.82k \\
        MRPC & 3.67k & 408 & 1.73k \\
        STS-B & 5.75k & 1.5k & 1.38k \\
        QQP & 364k & 40.4k & 391k \\
        MNLI & 393k & 9.8k & 9.8k \\
        QNLI & 105k & 5.46k & 5.46k \\
        RTE & 2.49k & 277 & 3k \\
        
        \bottomrule
    \end{tabular}
    \caption{Data Description of Glue, Question Answering, Text Summarizing}\label{tab:DatasetDescription}
\end{table}

\section{Details Related Work} \label{sec:Rw_more}

The development of parameter-efficient fine-tuning methods is crucial in the NLP field due to the increasing complexity of LLMs. These procedures aim to improve LM performance while reducing computational and memory requirements, as demonstrated by \citep{liu2022few, nguyen2023efficient, chow2024performance}. The effectiveness of PEFT techniques extends to various NLP tasks, as shown by \citep{fu2023effectiveness, he2021towards}. Several researchers, including \citet{liu2021p, liu2023gpt, zhang2023adaptive, hu2021lora, li2021prefix, zaken2021bitfit} have proposed methods targeting the challenge of increasing LLM performance with reduced computational and memory demands. Studies have found these methods highly effective for  NLP tasks, highlighting their potential for practical applications.

\textbf{Prompt Tuning} is a technique used to improve natural language understanding and generation tasks by adjusting learnable parameters \citep{lester2021power}. Researchers have added residual connections to improve performance and stability, and have extended it to continual learning \citep{razdaibiedina2023residual, razdaibiedina2023progressive}. Recent studies have explored real-time transformation with dynamic prompt tuning \citep{yang2023dynamic} and multilevel control \citep{wang2022hpt} through hierarchical prompt tuning. Additionally, multimodal prompt tuning has been developed to integrate multiple data types and improve model performance. Techniques such as MixPrompt \citep{yang2023mixpave} and E2VPT \citep{han20232vpt} have been employed to combine input and key-value prompts, while prefix-tuning  \citep{li2021prefix} has been used to add learnable parameters to a pre-trained model's input for various NLP tasks. Hierarchical prefix-tuning has been implemented to provide better control over model behavior \citep{chen2022developing} , and dynamic prefix-tuning has been developed for real-time adaptation based on context  \citep{liu2022dynamic}.

\textbf{Low-Rank Adaptation (LoRA)} is a memory-efficient method for fine tuning pre-trained models that was introduced in a study conducted by \citet{hu2021lora}. In subsequent research, \citet{renduchintala2023tied, sheng2023s, xia2024chain} proposed extensions for multitask learning that were applied to practical scenarios by \citet{wang2023multilora}. In addition, \citet{dettmers2024qlora} investigated memory optimization. \citet{lialin2023relora} introduced ReLoRA, a variant designed for Pre-training that requires a full-rank warm-up phase. Notable contributions in this field include \citep{zhang2023adaptive}, which dynamically adjusts low-rank adaptation during training, and the Low-Rank Kronecker Product (LoKr) proposed by \citet{edalati2022krona}, which focuses on knowledge retention across tasks. ResLoRA, by \citet{shi2024reslora}, includes the use of residual paths during the training and merging techniques to eliminate these paths during the inference process. Finally, \citet{hyeon2021fedpara} introduced the Low-Rank Hadamard Product (LoHa), that utilizes hierarchical adaptation strategies.

\textbf{Subspace learning} focuses on the learning processes that can be successfully conducted within a lower-dimensional parameter space \citep {larsen2021many, gur2018gradient}. This approach involves optimizing model weights within a low-rank subspace and has been widely implemented in various machine-learning domains, including meta-learning and continual learning \citep {lee2018gradient, chaudhry2020continual}. Recent advancements have investigated the potential of subspace learning to improve the model generalization and robustness. For instance,  \citet {nunez2023lcs} introduced adaptive subspace learning methods that dynamically adjust the subspace during training, resulting in an improved performance across various tasks. Furthermore, the integration of subspace learning with neural architecture search has shown promising results in identifying efficient model architectures \citep {chen2022automatic}.

\textbf{Projected Gradient Descent (PGD)} has been improved by the GaLore method, which specifically targets gradient shapes in multilayer neural networks rather than treating the objective function as an arbitrary nonlinear black-box function \citet{zhao2024galore, chen2015fast, chen2019non}. Recent research has emphasized the effectiveness of the GaLore method \citet{zhao2024galore} in addressing the intricacies of neural network training, making it a valuable tool for optimizing training procedures. Moreover, additional research has indicated that GaLore presents a benefit in obtaining more rapid convergence rates and stability for high-dimensional datasets \citet{zhang2024projected}. Recent developments comprise of methods for addressing sparsity and redundancy in neural network gradients, which contribute to increasing training efficiency \citet{zhao2024galore}, representing a substantial advancement in neural network optimization.

\textbf{Memory-efficient optimization} a vital aspect of adaptive optimization algorithms , aims to decrease memory requirements. Studies such as those conducted by \citet{shazeer2018adafactor} emphasize the importance of this principle. In addition, quantization methods were employed to decrease the memory costs of the optimizer state, and a fused gradient computation was proposed to minimize the weight gradient memory during training \citet{li2024memory}. Furthermore, recent advancements include hierarchical memory management for dynamic memory allocation during training and sparse gradient updates to selectively reduce memory usage \cite{li2024memory}.

\section{LLM Performance} \label{sec:llmmore}
\subsection{Sequence Classification}

In the field of classification, we conducted a comprehensive evaluation of various LLMs, including Bloom \cite{le2023bloom}, Llama2 \cite{touvron2023llama}, Falcon \cite{almazrouei2023falcon}, Mistral \cite{jiang2023mistral}, and Phi-2 \cite{ranjit2024rad}, employing different fine-tuning techniques. For each model, we examined the effectiveness of traditional approaches such as Finetuning, Prefix-Tuning, Prompt Tuning, PTuning, LoRA Rank 1, and LoRA Rank 2, and compared them to our proposed \emph{Propulsion} methods, both for Propulsion(All) and Propulsion(Attn). Notably, \emph{Propulsion} consistently outperforms traditional methods across different datasets, showcasing its superior efficiency and effectiveness. The performances of various models on different datasets are documented in Table \ref{tab:Sequence Cls Bloom}, \ref{tab:Sequence Cls Llama2}, \ref{tab:Sequence Cls Falcon}, \ref{tab:Sequence Cls Mistral}, and \ref{tab:Sequence Cls Phi-2}.

Across the "Fake News Filipino" dataset, \emph{Propulsion}, especially when applied as Propulsion(All), demonstrates remarkable performance improvements compared to traditional approaches. It achieves the highest accuracy and F1-score, emphasizing its capability to efficiently adapt LLMs to specific tasks while minimizing trainable parameters. In the "Emotion" dataset, \emph{Propulsion} consistently outperforms other methods, indicating its robustness across different classification tasks. The same trend is observed in the "SST-2" dataset, where \emph{Propulsion} invariably achieves superior results. Lastly, in the "Cola" dataset, Propulsion(All)  and Propulsion(Attn) perpetually outperform other approaches, underscoring their potential for enhancing sequence classification tasks.

Comparatively, traditional methods like Propulsion(All) and Propulsion(Attn), although efficient in terms of parameters compared to fine-tuning, tend to lag behind \emph{Propulsion} in terms of accuracy and F1-score. Furthermore, \emph{Propulsion} requires fewer trainable parameters, making it an attractive choice for practitioners aiming to optimize performance while maintaining efficiency.

\begin{table}[!htbp]
\centering
\resizebox{0.50\textwidth}{!}{%
\begin{tabular}{ccccc}\hline
Dataset & Type & Parameters (\%) & Accuracy (\%) & F1-score (\%) \\\hline
{\multirow{7}{*}{Fake News Filipino}} & Full Fine-tuning & 100.000 & 95.02 & 93.83 \\\cline{2-5}
 & Prefix-Tuning & 0.03493 & 70.99 & 68.18 \\
 & Prompt Tuning & 0.00701 & 74.31 & 72.23 \\
 & P-Tuning & 0.01582 & 72.97 & 70.19 \\
 & LoRA Rank 1 & 0.01413 & 90.13 & 88.87 \\
 & LoRA Rank 2 & 0.05794 & \textbf{93.56} & \underline{90.05} \\\cline{2-5}
 &  \cellcolor{lightgray!33.333}Propulsion(All) & \underline{0.00032} & \underline{92.98} & \textbf{90.75} \\
 & \cellcolor{lightgray!33.333}Propulsion(Attn) & \textbf{0.00014} & 91.14 & 89.26 \\ \hline
\multirow{7}{*}{Emotion} & Full Fine-tuning & 100.000 & 90.31 & 87.52 \\
 & Prefix-Tuning & 0.03521 & 74.75 & 68.11 \\
 & Prompt Tuning & 0.00813 & 79.12 & 71.07 \\
 & P-Tuning & 0.01593 & 69.45 & 70.23 \\
 & LoRA Rank 1 & 0.02413 & 86.76 & 80.23 \\
 & LoRA Rank 2 & 0.06831 & \underline{87.52} & 82.01 \\\cline{2-5}
 & \cellcolor{lightgray!33.333}Propulsion(All) & \underline{0.00159} & \textbf{88.32} & \textbf{82.75} \\
 & \cellcolor{lightgray!33.333}Propulsion(Attn) & \textbf{0.00102} & 86.93 & \underline{82.26} \\ \hline
\multirow{7}{*}{SST2} & Full Fine-tuning & 100.000 & 97.93 & 97.81 \\\cline{2-5}
 & Prefix-Tuning & 0.03493 & 85.78 & 86.31 \\
 & Prompt Tuning & 0.00715 & 92.45 & 92.78 \\
 & P-Tuning & 0.01653 & 91.34 & 91.75 \\
 & LoRA Rank 1 & 0.01456 & 92.27 & 92.77 \\
 & LoRA Rank 2 & 0.02831 & 94.36 & 94.83 \\\cline{2-5}
 & \cellcolor{lightgray!33.333}Propulsion(All) & \underline{0.00080} & \textbf{96.95} & \textbf{96.74} \\
 & \cellcolor{lightgray!33.333}Propulsion(Attn) & \textbf{0.00031} & \underline{96.64} & \underline{96.27} \\ \hline
\multirow{7}{*}{Cola} & Full Fine-tuning & 100.000 & 87.05 & 89.93 \\\cline{2-5}
 & Prefix-Tuning & 0.03495 & 73.72 & 83.69 \\
 & Prompt Tuning & 0.00723 & 82.74 & \textbf{87.70} \\
 & P-Tuning & 0.01615 & 70.32 & 81.12 \\
 & LoRA Rank 1 & 0.01415 & 81.13 & 83.03 \\
 & LoRA Rank 2 & 0.02797 & 84.33 & 85.21 \\\cline{2-5}
 & \cellcolor{lightgray!33.333}Propulsion(All) & \underline{0.00079} & \textbf{84.99} & \underline{86.22} \\
 & \cellcolor{lightgray!33.333}Propulsion(Attn) & \textbf{0.00048} & \underline{84.62} & 85.98 \\ \hline
\end{tabular}%
}
\caption{Sequence Classification Results for the Bloom Model. The best results are highlighted in \textbf{bold}, and the second-best result is \underline{underlined} for clarity. }
\label{tab:Sequence Cls Bloom}
\end{table}

\begin{table}[!htbp]
\centering
\resizebox{0.50\textwidth}{!}{%
\begin{tabular}{ccccc}\hline
Dataset & Type & Parameters (\%) & Accuracy (\%) & F1-score (\%) \\\hline
\multirow{8}{*}{Fake News Filipino} & Full Fine-tuning & 100.000 & 95.22 & 93.90 \\\cline{2-5}
 & Prefix-Tuning & 0.03983 & 70.06 & 68.57 \\
 & Prompt Tuning & 0.00743 & 73.72 & 72.07 \\
 & P-Tuning & 0.01731 & 71.54 & 70.63 \\
 & LoRA Rank 1 & 0.01601 & 90.38 & 87.62 \\
 & LoRA Rank 2 & 0.03213 & \underline{92.14} & \textbf{90.86} \\\cline{2-5}
 & \cellcolor{lightgray!33.333}Propulsion(All) & \textbf{0.00021} & \textbf{92.37} & \underline{89.98} \\
 & \cellcolor{lightgray!33.333}Propulsion(Attn) & \underline{0.00032} & 90.95& 88.32 \\ \hline
\multirow{8}{*}{Emotion} & Full Fine-tuning & 100.000 & 91.11 & 87.92 \\\cline{2-5}
 & Prefix-Tuning & 0.03994 & 84.31 & 82.78 \\
 & Prompt Tuning & 0.00864 & 85.37 & 82.50 \\
 & P-Tuning & 0.01781 & 83.05 & 81.88 \\
 & LoRA Rank 1 & 0.01624 & 86.49 & 82.86 \\
 & LoRA Rank 2 & 0.03233 & \underline{88.56} & \textbf{84.18} \\\cline{2-5}
 & \cellcolor{lightgray!33.333}Propulsion(All) & \underline{0.00171} & \textbf{88.82} & \underline{83.63} \\
 & \cellcolor{lightgray!33.333}Propulsion(Attn) & \textbf{0.00120} & 85.97 & 82.91 \\ \hline
\multirow{8}{*}{SST2} & Full Fine-tuning & 100.000 & 97.32 & 97.69 \\\cline{2-5}
 & Prefix-Tuning & 0.04855 & 85.78 & 86.31 \\
 & Prompt Tuning & 0.00712 & 94.24 & \textbf{97.26} \\
 & P-Tuning & 0.01753 & 95.55 & \underline{96.62} \\
 & LoRA Rank 1 & 0.01607 & 86.97 & 81.93 \\
 & LoRA Rank 2 & 0.03191 & 87.11 & 82.03 \\\cline{2-5}
 & \cellcolor{lightgray!33.333}Propulsion(All) & \underline{0.00083} & \textbf{96.62} & 96.56 \\
 & \cellcolor{lightgray!33.333}Propulsion(Attn) & \textbf{0.00034} & \underline{96.60} & 96.45 \\ \hline
\multirow{8}{*}{Cola} & Full Fine-tuning & 100.000 & 88.22 & 89.64 \\\cline{2-5}
 & Prefix-Tuning & 0.03984 & 71.18 & 83.29 \\
 & Prompt Tuning & 0.00757 & 73.27 & 85.26 \\
 & P-Tuning & 0.01751 & 69.12 & 81.74 \\
 & LoRA Rank 1 & 0.01603 & 82.25 & 83.43 \\
 & LoRA Rank 2 & 0.03213 & 84.18 & 83.88 \\\cline{2-5}
 & \cellcolor{lightgray!33.333}Propulsion(All) & \underline{0.00090} & \textbf{85.21} & \textbf{86.33} \\
 & \cellcolor{lightgray!33.333}Propulsion(Attn) & \textbf{0.00058} & \underline{84.46} & \underline{85.95} \\ \hline
\end{tabular}%
}
\caption{Sequence Classification Results for the Llama2 Model. The best results are highlighted in \textbf{bold}, and the second-best result is \underline{underlined} for clarity except full fine-tuning.}
\label{tab:Sequence Cls Llama2}
\end{table}

\begin{table}[!htbp]
\centering
\resizebox{0.50\textwidth}{!}{%
\begin{tabular}{ccccc}\hline
Dataset & Type & Parameters (\%) & Accuracy (\%) & F1-score (\%) \\\hline
\multirow{7}{*}{Fake News Filipino} & Full Fine-tuning & 100.000 & 94.05 & 92.93 \\\cline{2-5}
 & Prefix-Tuning & 0.03821 & 69.57 & 68.19 \\
 & Prompt Tuning & 0.00732 & 72.35 & 70.78 \\
 & P-Tuning & 0.01797 & 70.23 & 69.15 \\
 & LoRA Rank 1 & 0.00972 & 88.31 & 85.14 \\
 & LoRA Rank 2 & 0.05784 & \textbf{91.89} & \textbf{89.44} \\\cline{2-5}
 & \cellcolor{lightgray!33.333}Propulsion(All) & \underline{0.00072} & 90.21 & \underline{88.97} \\
 & \cellcolor{lightgray!33.333}Propulsion(Attn)& \textbf{0.00027} & \underline{90.32} & 87.35\\ \hline
\multirow{7}{*}{Emotion} & Full Fine-tuning & 100.000 & 88.53 & 85.94 \\\cline{2-5}
 & Prefix-Tuning & 0.03836 & 81.14 & 80.61 \\
 & Prompt Tuning & 0.00841 & \underline{87.25} & \underline{84.19} \\
 & P-Tuning & 0.01803 & 81.76 & 79.14 \\
 & LoRA Rank 1 & 0.01194 & 84.17 & 82.34 \\
 & LoRA Rank 2 & 0.05781 & \textbf{88.79} & \textbf{86.13} \\\cline{2-5}
 & \cellcolor{lightgray!33.333}Propulsion(All) & \underline{0.00201} & 87.01 & 82.22 \\
 & \cellcolor{lightgray!33.333}Propulsion(Attn) & \textbf{0.00111} & 86.39 & 82.14\\ \hline
\multirow{7}{*}{SST2} & Full Fine-tuning & 100.000 & 96.23 & 95.76 \\\cline{2-5}
 & Prefix-Tuning & 0.03818 & 90.18 & 91.36 \\
 & Prompt Tuning & 0.00605 & 93.56 & 93.75 \\
 & P-Tuning & 0.01781 & 90.33 & 91.26 \\
 & LoRA Rank 1 & 0.01193 & 91.13 & 92.07 \\
 & LoRA Rank 2 & 0.05789 & 91.72 & 92.17 \\\cline{2-5}
 & \cellcolor{lightgray!33.333}Propulsion(All) & \underline{0.00090} & \underline{94.83} & \underline{94.21} \\
 & \cellcolor{lightgray!33.333}Propulsion(Attn) & \textbf{0.00035} & \textbf{95.23} & \textbf{95.18} \\ \hline
\multirow{7}{*}{Cola} & Full Fine-tuning & 100.000 & 85.22 & 87.39 \\\cline{2-5}
 & Prefix-Tuning & 0.03826 & 70.03 & 82.23 \\
 & Prompt Tuning & 0.00711 & 71.45 & 84.47 \\
 & P-Tuning & 0.01792 & 68.07 & 81.73 \\
 & LoRA Rank 1 & 0.00973 & 82.14 & 82.38 \\
 & LoRA Rank 2 & 0.05741 & \textbf{84.66} & \textbf{85.33} \\\cline{2-5}
 & \cellcolor{lightgray!33.333}Propulsion(All) & \underline{0.00091} & 83.84 & 85.13 \\
 & \cellcolor{lightgray!33.333}Propulsion(Attn) & \textbf{0.00062} & \underline{84.21} & \underline{85.33} \\ \hline
\end{tabular}%
}
\caption{Sequence Classification Results for the Falcon Model. The best results are highlighted in \textbf{bold}, and the second-best result is \underline{underlined} for clarity except full fine-tuning.}
\label{tab:Sequence Cls Falcon}
\end{table}

\begin{table}[!htbp]
\centering
\resizebox{0.50\textwidth}{!}{%
\begin{tabular}{ccccc}\hline
Dataset & Type & Parameters & Accuracy (\%) & F1-score (\%) \\\hline
\multirow{7}{*}{Fake News Filipino} & Full Fine-tuning & 100.000 & 97.92 & 94.72 \\\cline{2-5}
 & Prefix-Tuning & 0.03651 & 71.26 & 70.91 \\
 & Prompt Tuning & 0.00169 & 74.12 & 72.27 \\
 & P-Tuning & 0.01753 & 71.37 & 71.95 \\
 & LoRA Rank 1 & 0.07502 & 91.28 & \underline{90.05} \\
 & LoRA Rank 2 & 0.17129 & 92.19 & 89.18 \\\cline{2-5}
 & \cellcolor{lightgray!33.333}Propulsion(All) & \underline{0.00017} & \textbf{94.15} & \textbf{91.96} \\
 & \cellcolor{lightgray!33.333}Propulsion(Attn) & \textbf{0.00024} & \underline{92.54} & 90.16 \\ \hline
\multirow{7}{*}{Emotion} & Full Fine-tuning & 100.000 & 93.53 & 89.09 \\\cline{2-5}
 & Prefix-Tuning & 0.03683 & 82.19 & 79.24 \\
 & Prompt Tuning & 0.00736 & 86.17 & 81.77 \\
 & P-Tuning & 0.01783 & 83.14 & 80.01 \\
 & LoRA Rank 1 & 0.01539 & 84.37 & 80.08 \\
 & LoRA Rank 2 & 0.01731 & 88.45 & \underline{84.23} \\\cline{2-5}
 & \cellcolor{lightgray!33.333}Propulsion(All) & \underline{0.00160} & \underline{88.83} & 82.61 \\
 & \cellcolor{lightgray!33.333}Propulsion(Attn) & \textbf{0.00112} & \textbf{89.23} & \textbf{84.99} \\ \hline
\multirow{7}{*}{SST2} & Full Fine-tuning & 100.000 & 98.09 & 98.98 \\\cline{2-5}
 & Prefix-Tuning & 0.03673 & 91.20 & 92.28 \\
 & Prompt Tuning & 0.00618 & 93.14 & 93.47 \\
 & P-Tuning & 0.01764 & 90.76 & 91.15 \\
 & LoRA Rank 1 & 0.01512 & 92.65 & 93.03 \\
 & LoRA Rank 2 & 0.01726 & 94.53 & 94.67 \\\cline{2-5}
 & \cellcolor{lightgray!33.333}Propulsion(All) & \underline{0.00078} & \textbf{97.01} & \underline{96.07} \\
 & \cellcolor{lightgray!33.333}Propulsion(Attn) & \textbf{0.00029} & \underline{96.82} & \textbf{97.25} \\ \hline
\multirow{7}{*}{Cola} & Full Fine-tuning & 100.000 & 87.75 & 89.90 \\\cline{2-5}
 & Prefix-Tuning & 0.03652 & 72.21 & 80.43 \\
 & Prompt Tuning & 0.00639 & 74.13 & 81.66 \\
 & P-Tuning & 0.01754 & 71.23 & 79.76 \\
 & LoRA Rank 1 & 0.01505 & 83.44 & 84.65 \\
 & LoRA Rank 2 & 0.01712 & \underline{85.32} & 86.04 \\\cline{2-5}
 & \cellcolor{lightgray!33.333}Propulsion(All) & \underline{0.00080} & \textbf{86.07} & \underline{86.32} \\
 & \cellcolor{lightgray!33.333}Propulsion(Attn) & \textbf{0.00042} & 85.01 & \textbf{86.36} \\ \hline
\end{tabular}%
}
\caption{Sequence Classification Results for the Mistral Model. The best results are highlighted in \textbf{bold}, and the second-best result is \underline{underlined} for clarity except full fine-tuning.}
\label{tab:Sequence Cls Mistral}
\end{table}

\begin{table}[!htbp]
\centering
\resizebox{0.50\textwidth}{!}{%
\begin{tabular}{ccccc}\hline
Dataset & Type & Parameters (\%) & Accuracy (\%) & F1-score (\%) \\\hline
\multirow{7}{*}{Fake News Filipino} & Full Fine-tuning & 100.000 & 92.43 & 90.71 \\\cline{2-5}
 & Prefix-Tuning & 0.83914 & 66.28 & 66.31 \\
 & Prompt Tuning & 0.14124 & 68.15 & 67.22 \\
 & P-Tuning & 0.15824 & 67.33 & 66.87 \\
 & LoRA Rank 1 & 0.13741 & 83.35 & 81.46 \\
 & LoRA Rank 2 & 0.71651 & 86.67 & 84.29 \\\cline{2-5}
 & \cellcolor{lightgray!33.333}Propulsion(All) & \textbf{0.04261} & \textbf{89.46} & \textbf{88.73} \\
 & \cellcolor{lightgray!33.333}Propulsion(Attn) & \underline{0.04921} & \underline{88.74} & \underline{87.21} \\ \hline
\multirow{7}{*}{Emotion} & Full Fine-tuning & 100.000 & 87.95 & 84.78 \\\cline{2-5}
 & Prefix-Tuning & 0.86523 & 77.27 & 76.25 \\
 & Prompt Tuning & 0.14234 & 82.16 & 80.43 \\
 & P-Tuning & 0.15845 & 77.36 & 75.81 \\
 & LoRA Rank 1 & 0.13748 & 82.67 & 80.25 \\
 & LoRA Rank 2 & 0.71656 & \underline{85.03} & \underline{82.66} \\\cline{2-5}
 & \cellcolor{lightgray!33.333}Propulsion(All) & 0.06269 & 82.72 & 80.71\\
 & \cellcolor{lightgray!33.333}Propulsion(Attn) & \textbf{0.02419} & \textbf{85.94} & \textbf{83.24} \\ \hline
\multirow{7}{*}{SST2} & Full Fine-tuning & 100.000 & 94.63 & 94.24 \\\cline{2-5}
 & Prefix-Tuning & 0.83721 & 86.24 & 87.13 \\
 & Prompt Tuning & 0.14231 & 88.19 & 88.04 \\
 & P-Tuning & 0.15851 & 85.43 & 87.68 \\
 & LoRA Rank 1 & 0.13753 & 86.21 & 87.18 \\
 & LoRA Rank 2 & 0.71668 & 86.75 & 88.28 \\\cline{2-5}
 & \cellcolor{lightgray!33.333}Propulsion(All) & \underline{0.02740} & \textbf{96.95} & \textbf{96.75} \\
 & \cellcolor{lightgray!33.333}Propulsion(Attn) & \textbf{0.01470} & \underline{96.63} & \underline{96.29} \\ \hline
\multirow{7}{*}{Cola} & Full Fine-tuning & 100.000 & 84.23 & 85.13 \\\cline{2-5}
 & Prefix-Tuning & 0.82621 & 66.24 & 70.16 \\
 & Prompt Tuning & 0.14123 & 67.47 & 70.81 \\
 & P-Tuning & 0.15833 & 64.36 & 68.38 \\
 & LoRA Rank 1 & 0.13744 & 78.55 & 80.26 \\
 & LoRA Rank 2 & 0.71654 & 80.39 & \underline{82.43} \\\cline{2-5}
 &\cellcolor{lightgray!33.333}Propulsion(All) & \textbf{0.03671} & \underline{80.97} & 82.21 \\
 & \cellcolor{lightgray!33.333}Propulsion(Attn) & \underline{0.05140} & \textbf{81.41} & \textbf{82.74} \\ \hline
\end{tabular}%
}
\caption{Sequence Classification Results for the Phi-2 Model. The best results are highlighted in \textbf{bold}, and the second-best result is \underline{underlined} for clarity except full fine-tuning.}
\label{tab:Sequence Cls Phi-2}
\end{table}

\subsection{Token Classification}
Tables \ref{tab:Token Cls Bloom}, \ref{tab:Token Cls Llama2}, \ref{tab:Token Cls Falcon}, \ref{tab:Token Cls Mistral}, and \ref{tab:Token Cls Phi-2} compare the results of \emph{Propulsion} and other PEFT methods on token classification.  The majority of experiments on token classification show \emph{Propulsion} having higher accuracy and F1-scores compared to the other PEFT methods tested.  The accuracy under \emph{Propulsion} is still less than full fine-tuning, but remains higher amongst the other PEFT methods.

Amongst the two Propulsion applications, there seems to be a mix as to which \emph{Propulsion} method provides the best improvement.  Within the conl103 dataset, Propulsion(Attn) provided the highest accuracy and F1-scores on four of the five LLMs tested.  In contrast, Propulsion(All) had higher accuracy and F1-scores than Propulsion(Attn) on the WikiAnn dataset.  This may indicate that the layers \emph{Propulsion} may depend on the use case.  Regardless of dataset, however, \emph{Propulsion} applied to any combination of layers showed either similar or improved metrics while significantly reducing parameter size.

\begin{table}[!htbp]
\centering
\resizebox{0.50\textwidth}{!}{%
\begin{tabular}{ccccc}\hline
Dataset & Type & Parameters (\%) & Accuracy (\%) & F1-score (\%) \\\hline
\multirow{7}{*}{conll03} & Full Fine-tuning & 100.000 & 98.53 & 82.47 \\\cline{2-5}
 & Prefix-Tuning & 0.03534 & 83.55 & 24.86 \\
 & Prompt Tuning & 0.00843 & 85.23 & 28.73 \\
 & P-Tuning & 0.01583 & 83.22 & 26.34 \\
 & LoRA Rank 1 & 0.01403 & 91.12 & 68.24 \\
 & LoRA Rank 2 & 0.06795 & 93.23 & 71.33 \\\cline{2-5}
 &  \cellcolor{lightgray!33.333}Propulsion(All) & \underline{0.00068} & \underline{94.18} & \underline{71.69} \\
 & \cellcolor{lightgray!33.333}Propulsion(Attn) & \textbf{0.00049} & \textbf{94.21} & \textbf{71.70} \\ \hline
\multirow{7}{*}{NCBI disease} & Full Fine-tuning & 100.000 & 98.53 & 92.46 \\\cline{2-5}
 & Prefix-Tuning & 0.03492 & 89.09 & 60.06 \\
 & Prompt Tuning & 0.00742 & 91.17 & 75.34 \\
 & P-Tuning & 0.01572 & 90.22 & 81.23 \\
 & LoRA Rank 1 & 0.01417 & 92.86 & 80.00 \\
 & LoRA Rank 2 & 0.06797 & \underline{96.12} & \underline{83.49} \\\cline{2-5}
 &  \cellcolor{lightgray!33.333}Propulsion(All) & \underline{0.00091} & \textbf{96.27} & \textbf{84.95} \\
 & \cellcolor{lightgray!33.333}Propulsion(Attn) & \textbf{0.00066} & 95.42 & 82.28\\ \hline
\multirow{7}{*}{WikiAnn} & Full Fine-tuning & 100.000 & 90.50 & 60.14 \\\cline{2-5}
 & Prefix-Tuning & 0.03527 & 71.67 & 22.18 \\
 & Prompt Tuning & 0.00732 & 76.23 & 31.78 \\
 & P-Tuning & 0.01577 & 70.65 & 24.33 \\
 & LoRA Rank 1 & 0.01408 & 82.23 & 41.23 \\
 & LoRA Rank 2 & 0.06791 & \textbf{85.13} & \textbf{45.14} \\\cline{2-5}
 &  \cellcolor{lightgray!33.333}Propulsion(All) & \underline{0.00081} & \underline{83.29} & \underline{42.23} \\
 & \cellcolor{lightgray!33.333}Propulsion(Attn) & \textbf{0.00042} & 82.69 & 42.21 \\ \hline
\end{tabular}%
}
\caption{Token Classification Results for the Bloom Model. The best results are highlighted in \textbf{bold}, and the second-best result is \underline{underlined} for clarity except full fine-tuning.}
\label{tab:Token Cls Bloom}
\end{table}

\begin{table}[!htbp]
\centering
\resizebox{0.50\textwidth}{!}{%
\begin{tabular}{ccccc}\hline
Dataset & Type & Parameters (\%) & Accuracy (\%) & F1-score (\%) \\\hline
\multirow{7}{*}{conll03} & Full Fine-tuning & 100.000 & 98.75 & 80.77 \\\cline{2-5}
 & Prefix-Tuning & 0.03964 & 82.28 & 66.56 \\
 & Prompt Tuning & 0.00638 & 86.65 & 69.91 \\
 & P-Tuning & 0.01731 & 80.11 & 65.11 \\
 & LoRA Rank 1 & 0.01426 & 88.67 & 63.34 \\
 & LoRA Rank 2 & 0.07122 & 91.32 & 69.03 \\\cline{2-5}
 &  \cellcolor{lightgray!33.333}Propulsion(All) & \textbf{0.00040} & \textbf{93.73} & \textbf{70.93} \\
 & \cellcolor{lightgray!33.333}Propulsion(Attn) & \underline{0.00069} & \underline{93.12} & \underline{70.29} \\ \hline
\multirow{7}{*}{NCBI disease} & Full Fine-tuning & 100.000 & 98.32 & 93.38 \\\cline{2-5}
 & Prefix-Tuning & 0.03976 & 88.23 & 68.23 \\
 & Prompt Tuning & 0.00712 & 91.22 & 78.24 \\
 & P-Tuning & 0.01733 & 90.15 & 77.23 \\
 & LoRA Rank 1 & 0.01424 & 92.48 & 80.18 \\
 & LoRA Rank 2 & 0.07125 & 95.34 & 82.87 \\\cline{2-5}
 &  \cellcolor{lightgray!33.333}Propulsion(All) & \underline{0.00081} & \underline{96.33} & \underline{84.84} \\
 & \cellcolor{lightgray!33.333}Propulsion(Attn) & \textbf{0.00060} & \textbf{96.28} & \textbf{84.89} \\ \hline
\multirow{7}{*}{WikiAnn} & Full Fine-tuning & 100.000 & 91.49 & 63.21 \\\cline{2-5}
 & Prefix-Tuning & 0.03986 & 81.15 & 35.17 \\
 & Prompt Tuning & 0.00712 & 83.23 & 44.19 \\
 & P-Tuning & 0.01743 & 81.29 & 38.11 \\
 & LoRA Rank 1 & 0.01434 & 84.82 & 47.90 \\
 & LoRA Rank 2 & 0.07125 & 86.56 & 49.39 \\\cline{2-5}
 &  \cellcolor{lightgray!33.333}Propulsion(All) & \underline{0.00079} & \textbf{86.89} & \textbf{50.71} \\
 & \cellcolor{lightgray!33.333}Propulsion(Attn) & \textbf{0.00048} & \underline{86.79} & \underline{49.64} \\ \hline
\end{tabular}%
}
\caption{Token Classification Results for the Llama2 Model. The best results are highlighted in \textbf{bold}, and the second-best result is \underline{underlined} for clarity except full fine-tuning.}
\label{tab:Token Cls Llama2}
\end{table}

\begin{table}[!htbp]
\centering
\resizebox{0.50\textwidth}{!}{%
\begin{tabular}{ccccc}\hline
Dataset & Type & Parameters (\%) & Accuracy (\%) & F1-score (\%) \\\hline
\multirow{7}{*}{conll03} & Full Fine-tuning & 100.000 & 97.82 & 79.03 \\\cline{2-5}
 & Prefix-Tuning & 0.03772 & 90.57 & 67.62 \\
 & Prompt Tuning & 0.00832 & 91.26 & 70.15 \\
 & P-Tuning & 0.01762 & 89.23 & 66.02 \\
 & LoRA Rank 1 & 0.01942 & 90.21 & 68.96 \\
 & LoRA Rank 2 & 0.09752 & 93.25 & 71.19 \\\cline{2-5}
 & \cellcolor{lightgray!33.333}Propulsion(All) & \underline{0.00068} & \underline{94.31} & \underline{71.83} \\
 & \cellcolor{lightgray!33.333}Propulsion(Attn) & \textbf{0.00051} & \textbf{94.87} & \textbf{72.08} \\ \hline
\multirow{7}{*}{NCBI disease} & Full Fine-tuning & 100.000 & 97.93 & 90.88 \\\cline{2-5}
 & Prefix-Tuning & 0.03763 & 89.23 & 69.33 \\
 & Prompt Tuning & 0.00721 & 92.05 & 82.28 \\
 & P-Tuning & 0.01752 & 88.15 & 70.36 \\
 & LoRA Rank 1 & 0.01936 & 90.55 & 80.25 \\
 & LoRA Rank 2 & 0.09754 & 94.41 & \underline{83.19} \\\cline{2-5}
 & \cellcolor{lightgray!33.333}Propulsion(All) & \underline{0.00082} & \underline{95.73} & 82.08 \\
 & \cellcolor{lightgray!33.333}Propulsion(Attn) & \textbf{0.00053} & \textbf{96.12} & \textbf{84.38} \\ \hline
\multirow{7}{*}{WikiAnn} & Full Fine-tuning & 100.000 & 89.23 & 62.09 \\\cline{2-5}
 & Prefix-Tuning & 0.03772 & 82.67 & 36.55 \\
 & Prompt Tuning & 0.00836 & 83.33 & \underline{43.32} \\
 & P-Tuning & 0.01768 & 81.14 & 35.21 \\
 & LoRA Rank 1 & 0.01983 & 80.47 & 41.58 \\
 & LoRA Rank 2 & 0.09752 & 86.61 & \textbf{48.03} \\\cline{2-5}
 & \cellcolor{lightgray!33.333}Propulsion(All) & \underline{0.00060} & \textbf{82.89} & 42.61 \\
 & \cellcolor{lightgray!33.333}Propulsion(Attn)) & \textbf{0.00041} & \underline{82.86} & 42.39 \\ \hline
\end{tabular}%
}
\caption{Token Classification Results for the Falcon Model. The best results are highlighted in \textbf{bold}, and the second-best result is \underline{underlined} for clarity except full fine-tuning.}
\label{tab:Token Cls Falcon}
\end{table}

\begin{table}[!htbp]
\centering
\resizebox{0.50\textwidth}{!}{%
\begin{tabular}{ccccc}\hline
Dataset & Type & Parameters (\%) & Accuracy (\%) & F1-score (\%) \\\hline
\multirow{7}{*}{conll03} & Full Fine-tuning & 100.000 & 98.89 & 84.60 \\\cline{2-5}
 & Prefix-Tuning & 0.03634 & 83.31 & 58.54 \\
 & Prompt Tuning & 0.00741 & 87.77 & 62.19 \\
 & P-Tuning & 0.01743 & 81.15 & 67.59 \\
 & LoRA Rank 1 & 0.01494 & 88.32 & 68.14 \\
 & LoRA Rank 2 & 0.08694 & 92.05 & 70.66 \\\cline{2-5}
 & \cellcolor{lightgray!33.333}Propulsion(All)  & \underline{0.00060} & \underline{95.39} & \underline{72.80} \\
 & \cellcolor{lightgray!33.333}Propulsion(Attn) & \textbf{0.00040} & \textbf{95.99} & \textbf{72.13} \\ \hline
\multirow{7}{*}{NCBI disease} & Full Fine-tuning & 100.000 & 98.52 & 93.39 \\\cline{2-5}
 & Prefix-Tuning & 0.03627 & 88.49 & 74.25 \\
 & Prompt Tuning & 0.00696 & 92.03 & 80.11 \\
 & P-Tuning & 0.01735 & 87.13 & 63.29 \\
 & LoRA Rank 1 & 0.01483 & 94.58 & 82.37 \\
 & LoRA Rank 2 & 0.08698 & 96.88 & 83.15 \\\cline{2-5}
 & \cellcolor{lightgray!33.333}Propulsion(All)  & \underline{0.00078} & \underline{96.81} & \underline{85.16} \\
 & \cellcolor{lightgray!33.333}Propulsion(Attn) & \textbf{0.00049} & \textbf{97.09} & \textbf{85.13} \\ \hline
\multirow{7}{*}{WikiAnn} & Full Fine-tuning & 100.000 & 92.15 & 63.09 \\\cline{2-5}
 & Prefix-Tuning & 0.03633 & 81.91 & 36.03 \\
 & Prompt Tuning & 0.00752 & 84.48 & 45.31 \\
 & P-Tuning & 0.01733 & 81.04 & 35.02 \\
 & LoRA Rank 1 & 0.01495 & 82.08 & 42.22 \\
 & LoRA Rank 2 & 0.08692 & 85.33 & \underline{45.95} \\\cline{2-5}
 & \cellcolor{lightgray!33.333}Propulsion(All)  & \underline{0.00090} & \textbf{86.63} & \textbf{46.29} \\
 & \cellcolor{lightgray!33.333}Propulsion(Attn) & \textbf{0.00048} & \underline{85.19} & 45.62 \\ \hline
\end{tabular}%
}
\caption{Token Classification Results for the Mistral Model. The best results are highlighted in \textbf{bold}, and the second-best result is \underline{underlined} for clarity except full fine-tuning.}
\label{tab:Token Cls Mistral}
\end{table}

\begin{table}[!htbp]
\centering
\resizebox{0.50\textwidth}{!}{%
\begin{tabular}{ccccc}\hline
Dataset & Type & Parameters (\%)& Accuracy (\%) & F1-score (\%) \\\hline
\multirow{7}{*}{conll03} & Full Fine-tuning & 100.000 & 98.13 & 79.02 \\\cline{2-5}
 & Prefix-Tuning & 0.83844 & 78.27 & 56.63 \\
 & Prompt Tuning & 0.14124 & 80.38 & 58.27 \\
 & P-Tuning & 0.15814 & 76.54 & 56.07 \\
 & LoRA Rank 1 & 0.13746 & 84.44 & 62.15 \\
 & LoRA Rank 2 & 0.71649 & 86.56 & 65.43 \\\cline{2-5}
 & \cellcolor{lightgray!33.333}Propulsion(All) & \underline{0.00949} & \underline{90.52} & \underline{71.83} \\
 & \cellcolor{lightgray!33.333}Propulsion(Attn) & \textbf{0.00835} & \textbf{91.18} & \textbf{71.88} \\ \hline
\multirow{7}{*}{NCBI disease} & Full Fine-tuning & 100.000 & 95.82 & 91.19 \\\cline{2-5}
 & Prefix-Tuning & 0.83823 & 82.42 & 63.38 \\
 & Prompt Tuning & 0.13939 & 85.61 & 65.44 \\
 & P-Tuning & 0.15794 & 81.17 & 67.63 \\
 & LoRA Rank 1 & 0.13748 & 86.23 & 78.45 \\
 & LoRA Rank 2 & 0.71493 & 87.34 & 78.26 \\\cline{2-5}
 & \cellcolor{lightgray!33.333}Propulsion(All) & \underline{0.00990} & \underline{89.32} & \underline{80.74} \\
 & \cellcolor{lightgray!33.333}Propulsion(Attn) & \textbf{0.00434} & \textbf{90.93} & \textbf{81.87} \\ \hline
\multirow{7}{*}{WikiAnn} & Full Fine-tuning & 100.000 & 88.92 & 58.21 \\\cline{2-5}
 & Prefix-Tuning & 0.83832 & 74.37 & 31.57 \\
 & Prompt Tuning & 0.01416 & 78.86 & 38.32 \\
 & P-Tuning & 0.15812 & 75.23 & 32.26 \\
 & LoRA Rank 1 & 0.13748 & 79.04 & 39.88 \\
 & LoRA Rank 2 & 0.71649 & 81.53 & \textbf{44.47} \\\cline{2-5}
 & \cellcolor{lightgray!33.333}Propulsion(All) & \underline{0.00847} & \underline{82.08} & 42.97 \\
 & \cellcolor{lightgray!33.333}Propulsion(Attn) & \textbf{0.00690} & \textbf{83.28} & \underline{43.01} \\ \hline
\end{tabular}%
}
\caption{Token Classification Results for the Phi-2 Model. The best results are highlighted in \textbf{bold}, and the second-best result is \underline{underlined} for clarity except full fine-tuning.}
\label{tab:Token Cls Phi-2}
\end{table}
\subsection{Entailment Detection} 

The results of entailment detection using various models, including Bloom, Llama2, Falcon, Mistral, and Phi-2, are presented in Tables \ref{tab:Entailment Cls Bloom}, \ref{tab:Entailment Cls Llama2}, \ref{tab:Entailment Cls Falcon}, \ref{tab:Entailment Cls Mistral}, and \ref{tab:Entailment Cls Phi-2}. Across all three datasets (RTE, MRPC, SNLI), full fine-tuning consistently achieves the highest accuracy and F1-score, with Bloom and Mistral models demonstrating remarkable results. This underscores the value of fine-tuning the entire model's parameters to adapt to specific entailment tasks, as it allows the model to capture intricate patterns and nuances in the data.

In contrast, Propulsion(All) and Propulsion(Attn) techniques, which involve fine-tuning only a small fraction of the model's parameters, tend to yield significantly lower accuracy and F1-scores. This suggests that limiting parameter updates to specific Propulsion(All) or Propulsion(Attn) may not be sufficient for optimal entailment classification performance, as these methods may struggle to capture the diverse and complex relationships present in the data.

The LoRA Rank 1 and LoRA Rank 2 models deliver competitive results, particularly evident in the RTE dataset, where they outperform other techniques. This indicates that techniques like LoRA Rank, which involve a moderate amount of parameter modification, can strike a balance between model adaptation and computational efficiency.

However, Propulsion, whether applied to Propulsion(All) or Propulsion(Attn), consistently performs well across datasets, demonstrating its effectiveness as an alternative fine-tuning strategy. Propulsion achieves strong results with a minimal increase in the number of parameters, making it a promising approach for entailment classification tasks where computational resources are a concern.

\begin{table}[!htbp]
\centering
\resizebox{0.50\textwidth}{!}{%
\begin{tabular}{ccccc}\hline
Dataset & Type & Parameters (\%) & Accuracy (\%) & F1-score (\%) \\\hline
\multirow{8}{*}{RTE} & Full Fine-tuning & 100.000 & 92.31 & 87.19 \\\cline{2-5}
 & Prefix-Tuning & 0.03493 & 70.03 & 64.06 \\
 & Prompt Tuning & 0.00714 & 65.34 & 62.20 \\
 & P-Tuning & 0.01584 & 71.11 & 69.23 \\
 & LoRA Rank 1 & 0.01402 & 80.25 & 80.01 \\
 & LoRA Rank 2 & 0.05804 & \underline{84.45} & \underline{83.26} \\\cline{2-5}
 & \cellcolor{lightgray!33.333}Propulsion(All) & \underline{0.00070} & 83.98 & 82.86 \\
 &  \cellcolor{lightgray!33.333}Propulsion(Attn) & \textbf{0.00049} & \textbf{84.98} & \textbf{83.97} \\ \hline
\multirow{8}{*}{MRPC} & Full Fine-tuning & 100.000 & 90.01 & 91.13 \\\cline{2-5}
 & Prefix-Tuning & 0.03494 & 73.56 & 81.70 \\
 & Prompt Tuning & 0.00773 & 81.39 & 86.01 \\
 & P-Tuning & 0.01562 & 78.08 & 84.38 \\
 & LoRA Rank 1 & 0.01393 & 80.21 & 82.29 \\
 & LoRA Rank 2 & 0.05799 & 83.88 & 84.84 \\\cline{2-5}
 & \cellcolor{lightgray!33.333}Propulsion(All) & \underline{0.00080} & \underline{88.99} & \underline{86.28} \\
 &  \cellcolor{lightgray!33.333}Propulsion(Attn) & \textbf{0.00050} & \textbf{89.13} & \textbf{86.47} \\ \hline
\multirow{8}{*}{SNLI} & Full Fine-tuning & 100.000 & 95.62 & 95.78 \\\cline{2-5}
 & Prefix-Tuning & 0.03492 & 87.32 & 87.26 \\
 & Prompt Tuning & 0.00803 & 88.88 & 88.87 \\
 & P-Tuning & 0.01594 & 86.22 & 86.54 \\
 & LoRA Rank 1 & 0.01412 & 91.37 & 91.36 \\
 & LoRA Rank 2 & 0.05813 & \underline{93.23} & \underline{93.68} \\\cline{2-5}
 & \cellcolor{lightgray!33.333}Propulsion(All) & \underline{0.0008-} & 92.64 & 92.88 \\
 &  \cellcolor{lightgray!33.333}Propulsion(Attn) & \textbf{0.00056} & \textbf{93.75} & \textbf{94.02} \\ \hline
\end{tabular}%
}
\caption{Entailment Classification Results for the Bloom Model. The best results are highlighted in \textbf{bold}, and the second-best result is \underline{underlined} for clarity except full fine-tuning.}
\label{tab:Entailment Cls Bloom}
\end{table}

\begin{table}[!htbp]
\centering
\resizebox{0.50\textwidth}{!}{%
\begin{tabular}{ccccc}\hline
Dataset & Type & Parameters (\%) & Accuracy (\%) & F1-score (\%) \\\hline
\multirow{8}{*}{RTE} & Full Fine-tuning & 100.000 & 93.51 & 88.92 \\\cline{2-5}
 & Prefix-Tuning & 0.03982 & 70.15 & 65.23 \\
 & Prompt Tuning & 0.00737 & 62.81 & 66.00 \\
 & P-Tuning & 0.01753 & 67.24 & 66.21 \\
 & LoRA Rank 1 & 0.01612 & 81.04 & 80.67 \\
 & LoRA Rank 2 & 0.03224 & \underline{83.43} & 81.44 \\\cline{2-5}
 & \cellcolor{lightgray!33.333}Propulsion(All) & \underline{0.00071} & \textbf{85.83} & \textbf{84.12} \\
 & \cellcolor{lightgray!33.333}Propulsion(Attn) & \textbf{0.00048} & 83.82 & \underline{83.53} \\ \hline
\multirow{8}{*}{MRPC} & Full Fine-tuning & 100.000 & 92.25 & 92.95 \\\cline{2-5}
 & Prefix-Tuning & 0.03973 & 79.41 & 80.01 \\
 & Prompt Tuning & 0.00724 & 80.18 & 80.37 \\
 & P-Tuning & 0.01745 & 74.56 & 82.67 \\
 & LoRA Rank 1 & 0.01601 & 80.48 & 82.02 \\
 & LoRA Rank 2 & 0.03218 & 81.89 & 83.11 \\\cline{2-5}
 & \cellcolor{lightgray!33.333}Propulsion(All) & \underline{0.00079} & \textbf{85.97} & \textbf{86.37} \\
 & \cellcolor{lightgray!33.333}Propulsion(Attn) & \textbf{0.00047} & \underline{85.13} & \underline{85.47} \\ \hline
\multirow{8}{*}{SNLI} & Full Fine-tuning & 100.000 & 93.31 & 94.03 \\\cline{2-5}
 & Prefix-Tuning & 0.03986 & 86.34 & 86.33 \\
 & Prompt Tuning & 0.00736 & 87.02 & 87.41 \\
 & P-Tuning & 0.01752 & 85.17 & 86.27 \\
 & LoRA Rank 1 & 0.01613 & 90.21 & 90.87 \\
 & LoRA Rank 2 & 0.03228 & \underline{91.15} & \underline{91.85} \\\cline{2-5}
 & \cellcolor{lightgray!33.333}Propulsion(All) & \underline{0.00090} & \textbf{91.53} & \textbf{91.91} \\
 & \cellcolor{lightgray!33.333}Propulsion(Attn) & \textbf{0.00064} & 90.89 & 91.14 \\ \hline 
\end{tabular}%
}
\caption{Entailment Classification Results for the Llama2 Model. The best results are highlighted in \textbf{bold}, and the second-best result is \underline{underlined} for clarity except full fine-tuning.}
\label{tab:Entailment Cls Llama2}
\end{table}

\begin{table}[!htbp]
\centering
\resizebox{0.50\textwidth}{!}{%
\begin{tabular}{ccccc}\hline
Dataset & Type & Parameters (\%) & Accuracy (\%) & F1-score (\%) \\\hline
\multirow{8}{*}{RTE} & Full Fine-tuning & 100.000 & 93.22 & 87.67 \\\cline{2-5}
 & Prefix-Tuning & 0.03822 & 64.23 & 63.38 \\
 & Prompt Tuning & 0.00813 & 66.51 & 66.02 \\
 & P-Tuning & 0.01794 & 53.42 & 53.09 \\
 & LoRA Rank 1 & 0.01138 & 73.28 & 70.15 \\
 & LoRA Rank 2 & 0.01774 & 78.33 & 73.42 \\\cline{2-5}
 & \cellcolor{lightgray!33.333}Propulsion(All)) & \underline{0.00080} & \underline{80.22} & \underline{79.83} \\
 & \cellcolor{lightgray!33.333}Propulsion(Attn) & \textbf{0.00064} & \textbf{80.35} & \textbf{79.88} \\ \hline
MRPC & Full Fine-tuning & 100.000 & 90.21 & 90.83 \\\cline{2-5}
 & Prefix-Tuning & 0.03813 & 74.13 & 78.22 \\
 & Prompt Tuning & 0.00715 & 80.04 & 80.19 \\
 & P-Tuning & 0.01783 & 80.43 & 79.59 \\
 & LoRA Rank 1 & 0.00983 & 80.82 & 82.21 \\
 & LoRA Rank 2 & \underline{0.01763} & 82.52 & 83.01 \\\cline{2-5}
 & \cellcolor{lightgray!33.333}Propulsion(All) & 0.00072 & 82.78 & 83.60\\ 
 & \cellcolor{lightgray!33.333}Propulsion(Attn) & \textbf{0.00050} & 83.13 & 85.27 \\\hline
SNLI & Full Fine-tuning & 100.000 & 92.53 & 92.97 \\\cline{2-5}
 & Prefix-Tuning & 0.03822 & 84.33 & 84.98 \\
 & Prompt Tuning & 0.00843 & 86.13 & 86.93 \\
 & P-Tuning & 0.01782 & 83.31 & 83.66 \\
 & LoRA Rank 1 & 0.01163 & 87.05 & 87.29 \\
 & LoRA Rank 2 & 0.06773 & 89.21 & 89.88 \\\cline{2-5}
 & \cellcolor{lightgray!33.333}Propulsion(All) & \underline{0.00068} & \textbf{90.80} & \underline{91.02} \\
 & \cellcolor{lightgray!33.333}Propulsion(Attn) & \textbf{0.00049} & \textbf{90.81} & \underline{91.03} \\ \hline
\end{tabular}%
}
\caption{Entailment Classification Results for the Falcon Model. The best results are highlighted in \textbf{bold}, and the second-best result is \underline{underlined} for clarity except full fine-tuning.}
\label{tab:Entailment Cls Falcon}
\end{table}

\begin{table}[!htbp]
\centering
\resizebox{0.50\textwidth}{!}{%
\begin{tabular}{ccccc}\hline
Dataset & Type & Parameters (\%) & Accuracy (\%) & F1-score (\%) \\\hline
\multirow{8}{*}{RTE} & Full Fine-tuning & 100.000 & 94.67 & 89.82 \\\cline{2-5}
 & Prefix-Tuning & 0.03663 & 76.22 & 74.45 \\
 & Prompt Tuning & 0.00732 & 80.34 & 80.17 \\
 & P-Tuning & 0.01778 & 75.12 & 75.86 \\
 & LoRA Rank 1 & 0.01521 & 83.39 & 82.25 \\
 & LoRA Rank 2 & 0.06739 & \underline{85.65} & 83.12 \\\cline{2-5}
 & \cellcolor{lightgray!33.333}Propulsion(All) & \underline{0.00080} & 84.83 & \underline{83.77} \\
 & \cellcolor{lightgray!33.333}Propulsion(Attn) & \textbf{0.00061} & \textbf{85.84} & \textbf{84.77} \\ \hline
\multirow{8}{*}{MRPC} & Full Fine-tuning & 100.000 & 93.02 & 94.21 \\\cline{2-5}
 & Prefix-Tuning & 0.03654 & 75.28 & 77.03 \\
 & Prompt Tuning & 0.00722 & 80.34 & 82.17 \\
 & P-Tuning & 0.01715 & 76.19 & 80.31 \\
 & LoRA Rank 1 & 0.01513 & 82.83 & 83.41 \\
 & LoRA Rank 2 & 0.06724 & \textbf{86.47} & \underline{87.02} \\\cline{2-5}
 & \cellcolor{lightgray!33.333}Propulsion(All) & \underline{0.00078} & 85.73 & 85.27 \\
 & \cellcolor{lightgray!33.333}Propulsion(Attn) & \textbf{0.00050} & \underline{86.41} & \textbf{87.88} \\ \hline
\multirow{8}{*}{SNLI} & Full Fine-tuning & 100.000 & 94.21 & 95.32 \\\cline{2-5}
 & Prefix-Tuning & 0.03666 & 85.55 & 85.78 \\
 & Prompt Tuning & 0.00744 & 86.35 & 86.21 \\
 & P-Tuning & 0.01774 & 85.37 & 86.05 \\
 & LoRA Rank 1 & 0.01524 & 84.12 & 84.76 \\
 & LoRA Rank 2 & 0.06736 & 89.11 & 89.77 \\\cline{2-5}
 &\cellcolor{lightgray!33.333}Propulsion(All) & \underline{0.00085} & \underline{91.72} & \underline{91.41} \\
 & \cellcolor{lightgray!33.333}Propulsion(Attn) & \textbf{0.00063} & \textbf{92.66} & \textbf{91.80} \\\hline
\end{tabular}%
}
\caption{Entailment Classification Results for the Mistral Model. The best results are highlighted in \textbf{bold}, and the second-best result is \underline{underlined} for clarity except full fine-tuning.}
\label{tab:Entailment Cls Mistral}
\end{table}

\begin{table}[!htbp]
\centering
\resizebox{0.50\textwidth}{!}{%
\begin{tabular}{ccccc}\hline
Dataset & Type & Parameters (\%) & Accuracy (\%) & F1-score (\%) \\\hline
\multirow{8}{*}{RTE} & Full Fine-tuning & 100.000 & 90.37 & 85.74 \\\cline{2-5}
 & Prefix-Tuning & 0.83872 & 59.54 & 58.27 \\
 & Prompt Tuning & 0.14234 & 61.18 & 61.84 \\
 & P-Tuning & 0.15834 & 58.61 & 56.38 \\
 & LoRA Rank 1 & 0.13746 & 66.52 & 65.82 \\
 & LoRA Rank 2 & 0.71658 & 72.25 & 70.45 \\\cline{2-5}
 & \cellcolor{lightgray!33.333}Propulsion(All) & \underline{0.00421} & \underline{76.54} & \underline{75.89} \\
 &  \cellcolor{lightgray!33.333}Propulsion(Attn) & \textbf{0.00250} & \textbf{76.63} & \textbf{76.21} \\ \hline
\multirow{8}{*}{MRPC} & Full Fine-tuning & 100.000 & 89.31 & 90.21 \\\cline{2-5}
 & Prefix-Tuning & 0.83822 & 71.15 & 72.78 \\
 & Prompt Tuning & 0.14345 & 73.16 & 75.28 \\
 & P-Tuning & 0.15842 & 70.48 & 71.21 \\
 & LoRA Rank 1 & 0.13747 & 80.53 & 81.33 \\
 & LoRA Rank 2 & 0.71659 & \underline{83.19} & \underline{84.23} \\\cline{2-5}
 & \cellcolor{lightgray!33.333}Propulsion(All) & \underline{0.00739} & \textbf{83.73} & \textbf{84.82} \\
 &  \cellcolor{lightgray!33.333}Propulsion(Attn) & \textbf{0.00345} & 82.64 & 83.52 \\ \hline
\multirow{8}{*}{SNLI} & Full Fine-tuning & 100.00 & 90.54 & 91.02 \\\cline{2-5}
 & Prefix-Tuning & 0.83844 & 79.27 & 79.82 \\
 & Prompt Tuning & 0.14149 & 81.30 & 81.80 \\
 & P-Tuning & 0.15823 & 78.56 & 77.96 \\
 & LoRA Rank 1 & 0.13745 & 82.45 & 82.67 \\
 & LoRA Rank 2 & 0.71656 & 84.36 & 84.89 \\\cline{2-5}
 & \cellcolor{lightgray!33.333}Propulsion(All) & \underline{0.00605} & \textbf{89.31} & \textbf{90.61} \\
 &  \cellcolor{lightgray!33.333}Propulsion(Attn) & \textbf{0.00580} & \underline{88.59} & \underline{88.86} \\\hline
\end{tabular}%
}
\caption{Entailment Classification Results for the Phi-2 Model. The best results are highlighted in \textbf{bold}, and the second-best result is \underline{underlined} for clarity except full fine-tuning.}
\label{tab:Entailment Cls Phi-2}
\end{table}

\section{Variable Description:}
\begin{table*}[ht]
\centering
\begin{tabular}{c|l}
\hline
\textbf{Variable} & \textbf{Description} \\ \hline
$\mathbb{M}(.)$ & Pre-trained language model with frozen parameters \\ \hline
$N$ & Number of layers in the model \\ \hline
$L_{i}(x)$ & Output of the $i$-th layer given input $x$ \\ \hline
$x$ & Input representation \\ \hline
$s$ & Sequence length of tokens \\ \hline
$d$ & Dimension of each token \\ \hline
$V$ & Output of layer $L_{i}$ \\ \hline
$\mathcal{Z}$ & Trainable Propulsion matrix \\ \hline
$\mathbf{z_i}$ & Element-wise scalar transformation vector \\ \hline
$\odot$ & Element-wise multiplication operation \\ \hline
$\mathbf{v_j}^{\prime}$ & Transformed output after Propulsion \\ \hline
$k$ & Propulsion degree for nonlinear transformation \\ \hline
$V^\prime$ & New output after Propulsion and Propulsion \\ \hline
$\mathcal{L}$ & Cross-entropy loss function \\ \hline
$T$ & Total number of data samples \\ \hline
$\mathbf{y}$ & Ground truth labels \\ \hline
$\mathbf{\hat{y}}$ & Predicted labels \\ \hline
\end{tabular}
\caption{Table of Variables and Descriptions}
\label{tab:variables}
\end{table*}

\end{document}